\pdfoutput=1  
\documentclass{article}
\usepackage[accepted]{tmlr}
\def\month{10}
\def\year{2026}
\def\openreview{\url{https://openreview.net/forum?id=gp77UrXyjT}}
\newcommand{\artifactstatement}{The code, configs, and evaluation artifacts used for the
reported experiments are available at \url{https://github.com/haewoon/leadership-coordination-control}.}
\newcommand{\showacknowledgments}{}
\AtBeginDocument{\hypersetup{
  pdftitle={Leadership as Coordination Control: Behavioral Signatures and the Recovery-Advantage Boundary in Multi-Agent LLM Teams},
  pdfauthor={Haewoon Kwak},
  pdfsubject={Transactions on Machine Learning Research (2026)},
  pdfkeywords={multi-agent LLM teams, coordination control, leadership, team science, contingency theory, behavioral signatures, ablation}}}

\usepackage{microtype}
\usepackage{hyperref}
\usepackage{url}
\usepackage{graphicx}
\usepackage{booktabs}
\usepackage{amsmath}
\usepackage{enumitem}
\usepackage{xcolor}

\newcommand{\sigstar}{\textsuperscript{*}}

\title{Leadership as Coordination Control: Behavioral Signatures and the Recovery-Advantage Boundary in Multi-Agent LLM Teams}
\author{\name Haewoon Kwak \email hwkwak@iu.edu \\
\addr Indiana University Bloomington}

\begin{document}

\maketitle

\begin{abstract}
Team science holds that leadership is \emph{contingent}: it helps only under specific conditions, and capable, autonomous teams may need none at all. We ask the analogous question for multi-agent LLM teams: under what \emph{measurable} conditions does process-level coordination control add value, and do those conditions match what team science predicts? Answering it takes measurement, not just accuracy: we use \emph{behavioral signatures} (majority lock-in, exploration, recovery from an incorrect round-0 consensus) and \emph{per-action ablations}, clean because each controller is an explicit action set rather than a monolithic prompt.

We operationalize three classical leadership styles (transactional, transformational, situational) as controllers over a shared action vocabulary (\texttt{explore}, \texttt{revise}, \texttt{accept}, \texttt{synthesize}), holding the agent set and final aggregation fixed. Team science \citep{hackman2002leading,bass1985leadership,hersey1969management} supplies the substrate: Bass's two-component structure maps to the \texttt{accept} and \texttt{revise} actions, giving the transactional controller its decomposition. As a matched controller with the same actions but an \emph{arbitrary} rule recovers no better than majority voting, it is the theory-derived rule, not the vocabulary, that does the work.

Across four task regimes and three open-weight model families on a single backend, no controller dominates by accuracy, as the contingency view predicts. Against a \emph{shared} round-0 vote, generated once and reused across conditions, transactional control matches the vote on all 12 (model, regime) combinations to within $1.3$pp, and accuracy gains appear in only two of the 36 leadership entries, situational and transformational, both on the single \texttt{llama-4-scout} social combination, where the round-0 majority is unreliable. Against the stronger flat baseline, only situational still gains ($+8$pp). A \emph{recovery-advantage} account, tested with four boundary probes, says when a controller beats plain interaction: only where the round-0 majority is unreliable, the task is recoverable, and undirected interaction does not already repair it. Because these conditions map onto contingency theory (leadership substitutes, path-goal redundancy, and the situational readiness gap), a largely null accuracy result is what the theory predicts, not a failure of the controllers. We advance this readiness-gap boundary as an explanatory hypothesis rather than an established general principle. Its null side is corroborated across every other combination, including under held-out prediction, while its positive side is observed in a single combination. We read process-level coordination control as a contingency to be measured and theory-mapped, not a leaderboard to be topped.
\end{abstract}
\section{Introduction}
Large language model (LLM) agents are increasingly deployed in collaborative settings where multiple agents deliberate, critique one another, and produce a joint answer. Recent multi-agent systems (MAS) work has explored several engineering patterns that improve over single-agent inference: multi-agent debate \citep{du2024improving,liang2024encouraging,chan2024chateval}, role specialization and pipelined decomposition \citep{hong2024metagpt,wu2023autogen,li2023camel,chen2023agentverse}, and self-refinement or critic-then-resolve loops \citep{madaan2023selfrefine,shinn2023reflexion}. These patterns operate at the \emph{knowledge level}: they shape what agents reason about, how reasoning is decomposed across agents, or whether a candidate answer is refined before final aggregation. Coordination is then implicit in the choice of debate format, decomposition rule, or refinement loop. We ask a complementary, \emph{process-level} question: given a fixed agent set and a fixed knowledge-combination scheme, what control vocabulary determines how disagreement is raised, revised, accepted, or reopened across rounds, and how do different choices in that vocabulary shape collective behavior? Team science suggests the answer is conditional. Leadership there is not universally beneficial but \emph{contingent} on team state: subordinate ability and clear tasks can \emph{substitute} for it entirely \citep{kerr1978substitutes}, and a leader adds value only by supplying a function the team is not otherwise getting \citep{hackman1986leading}. The sharper question is therefore \emph{when}, under what measurable conditions, process-level control adds value at all, and whether those conditions match what team science predicts.

We propose two methodological primitives for studying this process-level question. First, \emph{behavioral signatures} (majority lock-in, exploration rate, recovery from incorrect round-0 consensus, and dissent preservation, the last reflected in the recovery rate; Section~\ref{sec:evaluation}) replace single-number accuracy as the primary measurement target. Final accuracy is informative but varies sharply across model and regime combinations and is sensitive to per-model stochasticity (Appendix~\ref{app:statistics}); behavioral signatures, especially majority lock-in, produce within-controller contrasts that reproduce even when absolute accuracy rankings shift. Second, \emph{per-action ablations} expose which components of a controller carry the effect in which regime. Because each controller is specified as a small, explicit set of control actions rather than as a single black-box prompt, removing or substituting one action while keeping the rest active is a clean operation. Prompt-based controllers do not afford equally clean component-level ablations, since the mapping between prompt text and behavioral component runs through the LLM's interpretation.

As a substantive case study, we operationalize three classical leadership styles (transactional, transformational, situational) as process-level controllers over a small interaction-level action space (such as \texttt{explore}, \texttt{revise}, \texttt{accept}, \texttt{synthesize}; see Section~\ref{sec:experimental_setup} for the full set). Team science \citep{hackman2002leading,salas2005teamwork,bass1985leadership,hersey1969management} supplies the principled substrate: a theoretical organization of process, cognitive, and affective coordination levels (we concentrate on the first), and component-decomposable action sets. Bass's two-component structure (contingent reward, management by exception) maps directly to the \texttt{accept} and \texttt{revise} actions within transactional control, and a parallel decomposition (broadcast goal, differentiated directives) applies to transformational control. The substrate supplies theory-grounded action decomposition. It does not add coordination mechanisms beyond the multi-agent debate, role specialization, and self-refinement already explored at the knowledge level. A controller decides not what an agent should think, but whether the team should reopen deliberation, reduce dissent, or close the round; the two layers are complementary and our process-level vocabulary composes with, rather than replaces, existing knowledge-level mechanisms.

We test these methods on four task regimes (closed-ended QA \citep{talmor2019commonsenseqa,geva2021did}, abductive ambiguity \citep{bhagavatula2020abductive}, social-norm ambiguity \citep{forbes2020social}, and a mixed-workload benchmark interleaving the three) across three open-weight model families (\texttt{gpt-oss-120b}, \texttt{llama-4-scout}, \texttt{gemma-4-31B-it}) served through a single inference backend, plus a second tier of four boundary probes (a cross-domain extension to MATH-500 Level~5 \citep{hendrycks2021measuring,lightman2024lets}, adversarial NLI \citep{nie2020adversarial}, Winogrande \citep{sakaguchi2021winogrande}, and a contested moral-judgment set \citep{lourie2021scruples}) that deliberately vary round-0 reliability and recoverability (Section~\ref{sec:boundary}). The findings answer the contingency question. No controller dominates by accuracy, as the contingency view predicts, and the conditions under which one helps are organized by a single measurable axis, the reliability of the independent round-0 majority (how often it is correct): a controller adds value only where that majority is unreliable, the task is recoverable (its incorrect majority can be repaired), and undirected interaction does not already repair it. We then map this measured boundary back to team-science contingency theory (Section~\ref{sec:theory_mapping}), where each region corresponds to a named construct: leadership substitutes, path-goal redundancy (undirected interaction already reaches the answer, leaving direction nothing to add), and the situational readiness gap (the team can improve, but only with direction).

Our contributions are:
\begin{itemize}[leftmargin=*,noitemsep]
    \item Two measurement primitives for process-level coordination control: \emph{behavioral signatures} as the primary scientific object replacing single-number accuracy, and \emph{per-action ablations} that decompose a controller by representing it as a small explicit action set. They produce within-controller and within-component contrasts, most robustly majority lock-in, that reproduce across the matrix even where accuracy rankings do not. A per-action ablation is diagnostic only where a component fires often enough to measure; where one is dormant, the methodology reveals that fact rather than producing a misleading null.
    \item Operationalization of three classical leadership theories as process-level controllers: the first substantive case study of the methodology, with team science supplying the principled substrate for component-decomposable action sets.
    \item A decomposition of each controller's gain into \emph{recovery} and \emph{breakage}: recovery repairs incorrect round-0 majorities and breakage corrupts correct ones, making the internal dynamics of multi-agent deliberation legible where a single accuracy number is silent. It reorganizes otherwise scattered controller-label results onto one measurable axis, the reliability of the independent round-0 majority, and isolates a \emph{recovery advantage} over plain interaction as the precise condition under which a controller earns its structure, met in exactly one combination (Section~\ref{sec:boundary}). Because the positive side of that boundary rests on this single combination, we present the boundary as an explanatory hypothesis whose null side is corroborated across the rest of the matrix (including under held-out prediction; Section~\ref{sec:predictive_analysis}).
    \item A mapping from the measured boundary back to leadership contingency theory (Section~\ref{sec:theory_mapping}): a reliable round-0 acts as a \emph{leadership substitute}, an unrecoverable task offers no latent ability to activate, undirected recovery makes control \emph{redundant} (path-goal), and the single region where control helps is the \emph{readiness gap} situational leadership predicts.
\end{itemize}

\paragraph{Terminology.}
Since several terms recur before their formal definitions (Sections~\ref{sec:method} and~\ref{sec:evaluation}), we explain them here for readers new to the team-science vocabulary. Of the three leadership styles, \emph{transactional} leadership manages by standards and corrective feedback on individual outputs; \emph{transformational} leadership aligns the team on a shared goal and encourages exploration around it; \emph{situational} leadership chooses its intervention from the team's current state, its \emph{readiness}, not from a fixed style. \emph{Flat} denotes undirected multi-round deliberation with no controller. On the measurement side, the \emph{round-0 majority} (or \emph{round-0 vote}) is the majority answer over the agents' independent first answers, taken before any interaction. \emph{Majority lock-in} is the fraction of episodes whose final answer equals that round-0 majority; \emph{exploration} is the fraction of episodes in which the controller issues an \texttt{explore} action; \emph{recovery} is the probability that the final answer is correct given that the round-0 majority was wrong, and \emph{breakage} is the reverse (a correct round-0 majority dragged to a wrong final answer). The \emph{recovery advantage} is a controller's recovery minus flat's on the same items, and the \emph{readiness gap} is the team state in which the team's independent answer is wrong but the team can reach the right answer when its deliberation is directed.

\section{Related Work}

\paragraph{Process- and knowledge-level coordination in multi-agent LLM systems.}
Recent multi-agent systems literature has explored several engineering patterns that improve over single-agent inference. \emph{Multi-agent debate} (MAD) variants \citep{du2024improving,liang2024encouraging,chan2024chateval} have agents independently propose, then iteratively critique and revise toward consensus, aggregating by majority vote or a final-round judgment. \emph{Role specialization and pipelined decomposition} frameworks \citep{hong2024metagpt,wu2023autogen,li2023camel,chen2023agentverse} assign distinct personas, sub-tasks, or pipeline positions to different agents and route communication accordingly. \emph{Self-refinement and critic-then-resolve loops} \citep{madaan2023selfrefine,shinn2023reflexion} use a single agent (or a critic head) to iteratively improve outputs through structured feedback. These mechanisms primarily operate at the \emph{knowledge level}: they shape \emph{what} each agent reasons about, \emph{which} agent contributes which subtask, or \emph{whether} a candidate answer is refined further before final aggregation; coordination is then implicit in the choice of debate format, decomposition rule, or aggregation procedure. Our work is complementary. We hold the agent set and the final aggregation step fixed and study a small \emph{process-level} action vocabulary (when to explore, revise, accept, or synthesize, with policy-specific extensions) that shapes round-to-round interaction dynamics rather than the substantive content of agent reasoning. We compare against a peer-to-peer MAD baseline directly in Appendix~\ref{app:mad_siturand}.

\paragraph{Behavioral measurement and component-level controller analysis.}
Most MAS evaluations report final accuracy, rounds, and token cost; trajectory-level behavioral metrics are reported irregularly and rarely treated as the primary measurement target \citep{du2024improving,liang2024encouraging,chan2024chateval,hong2024metagpt,madaan2023selfrefine}. Process metrics that do appear (turn counts, agreement rates, retry frequencies) are typically descriptive add-ons rather than controller-discriminating instruments. Two methodological gaps follow. First, behavioral measurement is sharpest when paired with controllers whose action sets are explicit: signatures like lock-in and recovery contrast most cleanly when one knows which control action ``locks in'' or ``reopens.'' Second, component-level ablations of MAS controllers are uncommon because controllers are typically specified as monolithic prompts; ablating a prompt fragment does not cleanly ablate a behavioral component. We address both gaps by representing controllers as theory-derived explicit action sets, which makes behavioral signatures comparable across controllers and per-action ablations clean by construction. That majority voting is a strong baseline for multi-agent LLM systems is increasingly recognized; our contribution is not that observation but the measurement that localizes it: a shared-round-0 attribution and a recovery/breakage decomposition that identify which controllers add value over voting and over plain interaction, where, and by what mechanism. A matched arbitrary controller (the same action set assigned at random) recovers no better than majority voting (Appendix~\ref{app:random_action}), confirming that the theory-derived rule, not the vocabulary, is what the per-action ablations isolate.

\paragraph{Team science as a motivating frame.}
A broader line of team-science and organizational research argues that team performance is not determined by structure alone \citep{hackman2002leading,salas2005teamwork,cooke2015team,greer2018leading}. Diversity can broaden coverage while increasing friction; hierarchy can improve accountability while also suppressing information flow \citep{cox1991managing,horwitz2007effects,van2004diversity}. These findings motivate a design language for agent teams that goes beyond ``flat versus hierarchical'' or ``homogeneous versus diverse'' and instead asks what mechanisms regulate conflict, revision, coordination cost, and adaptation once a team is already assembled \citep{muralidharan2025lessons}.

\paragraph{Leadership theories used as controller specifications.}
We choose three leadership theories because together they span three distinct axes of the operational leadership literature. Transactional leadership emphasizes standards, monitoring, and contingent feedback \citep{burns1978leadership,bass1985leadership}; transformational leadership emphasizes shared goal alignment coupled with structured exploration \citep{bass1985leadership}; situational leadership treats the appropriate style as a function of team state rather than a fixed identity \citep{hersey1969management}. The triad spans \emph{focus} (correction vs.\ goal-framing), \emph{level} (individual contingent feedback vs.\ collective objective), and \emph{adaptivity} (fixed style vs.\ state-driven switching). Other styles (servant, authentic, inclusive) are useful at the value or relational level but do not introduce a new control axis at the operational level we model. We treat them either as extensions of the same control vocabulary (e.g., inclusive leadership as a dissent-preservation specialization, included as a probe in Appendix~\ref{app:focused_social_core}) or as scope for future work.

\paragraph{Leadership is contingent, not universal.}
Team science has long held that leadership effectiveness depends on the situation, not the style alone. Contingency \citep{fiedler1967theory} and path-goal \citep{house1971path} theories tie a leader's value to situational favorableness and to whether the leader supplies something the team lacks; \emph{functional} leadership \citep{hackman1986leading} frames the leader's job as doing ``whatever is not being adequately handled for group needs''; and \emph{leadership-substitutes} theory \citep{kerr1978substitutes} identifies subordinate ability, experience, and intrinsically clear tasks as conditions under which leadership is unnecessary or neutralized. Situational leadership \citep{hersey1969management} makes the dependence explicit, switching style with follower readiness. We adopt this contingency stance as our null hypothesis rather than assuming control helps: we expect process-level control to add value only in a restricted regime, and we use the measurement vocabulary above to identify which team states correspond to each predicted regime (Section~\ref{sec:theory_mapping}).

\section{Leadership as Coordination Control}
\label{sec:method}
This section specifies the process-level control framework that the methodology operates on, then instantiates it with three classical leadership styles as the first case study. Let a team of agents be denoted by $A = \{a_1, \dots, a_n\}$. At interaction step $t$, each agent produces an intermediate output $y_{i,t}$ conditioned on task input $x$, interaction memory $M_t$, and the current team state $S_t$. We define a process-level control policy
\[
\pi : (S_t, Y_t, M_t) \rightarrow U_t,
\]
where $Y_t = \{y_{1,t}, \dots, y_{n,t}\}$ and $U_t$ is a set of control actions such as \texttt{revise}, \texttt{accept}, \texttt{explore}, or \texttt{synthesize}. A controller acts on the interaction process rather than only on prompt context, and it can be evaluated through its behavioral consequences. The leadership case study below uses three classical styles (transactional, transformational, situational) to instantiate the policy $\pi$ with theory-grounded action sets; Section~\ref{sec:bass_two_pillar} adds the Bass two-component decomposition that the per-action ablations target.

\subsection{Leadership as Control}
All conditions share a common loop (Table~\ref{tab:controller_skeleton}). Agents first answer independently in round 0, then receive bounded memory plus the policy-conditioned control actions defined below. After each round, the team either synthesizes a final answer or continues until convergence or budget exhaustion. This keeps comparisons centered on controller behavior under a matched interaction scaffold rather than on changes in team structure, model choice, or execution budget.

\begin{table}[t]
\centering
\small
\fbox{%
\parbox{0.94\columnwidth}{%
\textbf{Controller Skeleton}\par
\texttt{Input: task $x$, team outputs $Y_0$, state $S_0$}\par
\texttt{for round $t=0,\dots,T$:}\par
\hspace*{1em}\texttt{assess disagreement / rationale conflict / early consensus}\par
\hspace*{1em}\texttt{if policy = transactional: revise or accept, then synthesize}\par
\hspace*{1em}\texttt{if policy = transformational: explore via differentiated roles, then synthesize}\par
\hspace*{1em}\texttt{if policy = situational: choose explore vs. converge from team state}\par
\hspace*{1em}\texttt{update memory $M_t$ and team state $S_t$}\par
\texttt{return final team answer}
}}
\caption{Compact view of leadership as control. The policies differ not in agent identity or prompt role alone, but in which control actions they trigger and when.}
\label{tab:controller_skeleton}
\end{table}

The control space $U_t$ contains a small set of interventions:
\begin{itemize}[leftmargin=*,noitemsep]
    \item \texttt{explore}: instruct an agent to pursue a distinct line of reasoning, evidence source, or decomposition strategy;
    \item \texttt{revise}: require an agent to update its current answer based on criticism or mismatch with the team objective;
    \item \texttt{accept}: mark an intermediate answer as good enough to enter the synthesis pool without further revision;
    \item \texttt{synthesize}: aggregate accepted intermediate outputs into a team-level answer;
    \item \texttt{justify}: request explicit explanation or evidence from an agent before the output can be accepted.
\end{itemize}
Policies differ in how often they invoke these actions, which is itself part of the evaluation. Appendix~\ref{app:policy_pseudocode} gives compact pseudocode for each policy so that the operational differences are easy to inspect.

\subsection{Policy Instantiations and Baselines}
\paragraph{Transactional leadership.}
The transactional controller implements explicit oversight and criterion-based feedback \citep{burns1978leadership,bass1985leadership}. For each agent:
\[
u_{i,t} = C(y_{i,t}, r_t),
\]
where $C$ is a critic function and $r_t$ is the current acceptance criterion. If $u_{i,t}=0$, the agent revises; if $u_{i,t}=1$, the output is accepted into the synthesis pool. In practice this produces a centralized revise--accept loop with strong convergence pressure.

\paragraph{Transformational leadership.}
The transformational controller implements shared goal framing plus structured exploration \citep{bass1985leadership}. The leader broadcasts a shared \emph{goal} $g_t$ (the \emph{broadcast goal}) while assigning differentiated exploration directives $d_{i,t}$:
\[
x'_{i,t} = f(x, g_t, d_{i,t}, M_t).
\]
The purpose is coordinated variation around the broadcast goal rather than unconstrained divergence. Operationally, the leader intervenes less through local correction and more through framing, complementary exploration roles, and delayed synthesis.

\paragraph{Situational leadership.}
Situational leadership is an adaptive controller \citep{hersey1969management} that switches between exploration-oriented and convergence-oriented behavior based on team state, not task identity. The switch uses fixed, low-degree-of-freedom thresholds rather than tuned parameters. Concretely, for a 3-agent team the controller opens one exploratory step before synthesis whenever the round's answers are not unanimous (a three-way split, or a 2--1 split) or are nominally unanimous but show genuine underlying disagreement, operationalized as at least three distinct rationale openings (first-sentence prefixes) together with conflict markers (e.g., ``however,'' ``contradict,'' ``unlikely'') from at least two agents. Otherwise it falls back to a transactional-style convergence step, and it closes the round once a 2-of-3 majority agrees (Appendix~\ref{app:policy_pseudocode} gives the full rule). This keeps the policy interpretable: its goal is not to invent a new specialist strategy, but to decide when the team should reopen deliberation and when it should close it.

\paragraph{Baselines.}
We compare these policies against two baselines.
\begin{itemize}[leftmargin=*,noitemsep]
    \item \textbf{Flat}: symmetric interaction and final aggregation by vote or simple synthesis.
    \item \textbf{Theory-free control}: a leader performs generic oversight using fixed critique loops or arbitration heuristics without theoretical grounding.
\end{itemize}

The theory-free control baseline separates theory-grounded leadership from generic intervention. It uses a simple generic oversight rule: if the team has not yet converged, all agents are asked to revise; otherwise the team synthesizes. Unlike transactional or transformational control, it has no explicit theory of why disagreement should be reduced or preserved.

\subsection{Bass Two-Component Decomposition}
\label{sec:bass_two_pillar}
The transactional and transformational controllers each correspond to a two-component structure in Bass's leadership theory \citep{bass1985leadership}, which we make explicit because it renders both controllers \emph{component-decomposable}. Transactional leadership pairs \emph{contingent reward} (acknowledge and accept behavior that meets the criterion) with \emph{management by exception} (intervene when behavior fails the criterion). In our control vocabulary, these map directly to the \texttt{accept} and \texttt{revise} actions, which lets us ablate the management-by-exception action by holding \texttt{accept} active and removing \texttt{revise} from the control space. Transformational leadership pairs \emph{inspirational motivation} (a shared, energizing collective objective) with \emph{individualized consideration} (differentiated attention to each follower's contribution). In our vocabulary, these map respectively to the broadcast goal $g_t$ and the per-agent directives $d_{i,t}$, and we ablate the broadcast-goal component by removing $g_t$ while keeping $d_{i,t}$. Per-component ablations on both controllers are revisited empirically in later sections, where they isolate which component carries the behavioral signature in each regime and which becomes inactive, or actively harmful, as the task gets harder.

\subsection{Implementation Assumptions}
We focus on 3-agent teams so that traces remain comparable and interpretable. Within each benchmark--model run, all policy conditions share the same base model, configured generation settings, and task-type round budgets; Appendix~\ref{app:execution_settings} lists the concrete settings used in the main and cross-model evaluations.

The policy rules were not tuned on a held-out set to maximize accuracy. Instead, we use simple, low-degree-of-freedom thresholds chosen to keep the controllers interpretable and faithful to their intended leadership styles. The thresholds are stated explicitly (the situational trigger above, and the full per-policy rules in Appendix~\ref{app:policy_pseudocode}) so that the operational difference between controllers does not hide in prompt text. Several rules follow directly from team structure, such as 2-of-3 majority convergence in transactional control, while others are deliberately conservative and only trigger extra exploration under clear signs of unstable consensus, rationale conflict, or explicit objection. This is a proof-of-concept design choice: our goal is to test whether leadership theories can be operationalized as measurable coordination mechanisms before optimizing the resulting controllers for peak task performance. A sensitivity analysis (Appendix~\ref{app:situational_sensitivity}) confirms that the situational controller's social-regime gain is robust to the conflict-marker lexicon and to the precise thresholds (it survives a lexicon swap and grows under threshold relaxation) and is localized to the 2--1-split reopening trigger rather than any tuned hyperparameter.

\paragraph{Operationalization, not unique mapping.}
The controllers above are operationalizations of their source theories rather than uniquely correct implementations. The released configs and code allow other groups to substitute alternative operationalizations (e.g., entropy-based exploration triggers, dissent-preserving inclusive variants) within the same control-action vocabulary. The scientific claim is not that any particular controller is the canonical embodiment of transactional or transformational leadership, but that this control-action vocabulary is rich enough to produce distinct, measurable behavioral signatures across model families.

\subsection{Cross-Round Majority Extraction}
\label{sec:cross_round_extraction}
Open-ended numeric tasks introduce an answer-extraction subproblem independent of controller logic: when an agent's final reasoning is long, a last-line single-pass extractor occasionally picks up a stray number from the trailing trace rather than the agent's actual \texttt{Final answer:} commitment. We use a cross-round majority extractor as the default finalize step in all open-ended numeric evaluations: it scans every agent-round \texttt{Final answer:} declaration, normalizes the extracted answers, and takes the majority across all rounds and agents, falling back to the single-pass extractor where no such declaration was produced. The fix is a \emph{measurement} improvement decoupled from controller logic. All conditions including \texttt{flat} benefit from it, with the leadership controllers gaining the largest fraction of the recovered accuracy. Cross-controller comparisons on open-ended numeric tasks therefore remain on a common basis.

\section{Experimental Scope}
\subsection{Task Regimes}
We evaluate leadership control in three base shared-information regimes, plus a mixed workload that interleaves all three (Section~\ref{sec:experimental_setup}), for four task regimes in total. In all cases, agents see the same task input and answer space but receive different reasoning profiles: short instruction prefixes that orient each of the three agents toward a distinct judgment strategy (e.g., counterexample-first, surface-match, plausibility-first; the per-regime prefixes are released with the configs). This isolates coordination under divergent judgment without introducing private evidence. We now describe the three base regimes, each posing a different coordination challenge.

\paragraph{Verification-heavy closed-ended QA.}
This regime is built from selected CommonsenseQA and StrategyQA items \citep{talmor2019commonsenseqa,geva2021did}. It measures whether explicit control improves reliability under tasks that reward careful checking and efficient convergence.

\paragraph{Abductive ambiguity.}
This regime uses AlphaNLI examples \citep{bhagavatula2020abductive} in which multiple hypotheses remain initially plausible. It tests whether leadership can avoid premature convergence when early superficial plausibility is misleading.

\paragraph{Social-norm ambiguity.}
This regime uses broad Social Chemistry items \citep{forbes2020social} drawn from \texttt{confessions}, \texttt{amitheasshole}, and \texttt{dearabby}, so that the evaluation reflects heterogeneous social judgments rather than one narrow norm domain. It tests whether leadership can manage that heterogeneity, especially when context-sensitive and clearly unacceptable judgments coexist.

\subsection{Selection Protocol and Reporting Layers}
Because the paper is regime-focused rather than leaderboard-focused, each benchmark family is reported through two layers: a \emph{broad layer} used as an anti-cherrypicking evaluation set, and a narrower \emph{curated layer} used when a cleaner mechanism readout is needed.

Broad layers are constructed by fixed rules rather than by policy outcome. Closed-ended uses a balanced CommonsenseQA+StrategyQA pool with only a small fixed exclusion list. AlphaNLI uses the stable reference pool directly at 100-scale. Social Chemistry keeps only \texttt{B/C} items with intermediate agreement from conflict-rich areas. Appendix~\ref{app:curation_protocol} and Table~\ref{tab:appendix_curation_summary} provide the regime-by-regime construction summary, exclusion logic, and source composition. In the main text, the important point is simpler: curated results sharpen mechanism visibility, while broad-layer results test whether the same pattern survives when the workload is widened.

\subsection{Experimental Setup}
\label{sec:experimental_setup}
The main comparison uses five conditions: (1) flat deliberation, (2) theory-free control, (3) transactional leadership, (4) transformational leadership, and (5) situational leadership; Section~\ref{sec:bass_empirical} adds two Bass component ablations (transactional accept-only, transformational directives-only).

We evaluate these conditions on three open-weight model families served through a single self-hosted inference backend: \texttt{gpt-oss-120b}, \texttt{llama-4-scout}, and \texttt{gemma-4-31B-it}.\footnote{Names are the checkpoint identifiers exposed by that backend and correspond to public open-weight releases: \texttt{gpt-oss-120b} (OpenAI's open-weight 120B reasoning model), \texttt{llama-4-scout} (Meta Llama~4 Scout), and \texttt{gemma-4-31B-it} (a Google Gemma release in the $\sim$30B class). We report the gateway identifiers verbatim for reproducibility; exact revisions are pinned in the released configs.} Using one backend for every model avoids the cross-backend heterogeneity that a mix of locally-quantized and API-served models would introduce, so that accuracy differences reflect model and controller behavior, not serving conditions. The three families also span a capability range, which matters because the controllers' effect depends on how reliable each model's independent reasoning is. The main matrix is 3 model families $\times$ 4 task regimes (closed-ended QA, abductive ambiguity, social-norm ambiguity, and a mixed workload), 12 (model, regime) combinations; the same three models are the testbed for the MATH-500 Level 5 extension (Section~\ref{sec:math500}).

\paragraph{Two evaluation tiers.}
We organize the evaluation into two tiers. The \emph{main matrix} (the three regimes above plus the mixed workload, across the three models) covers regimes in which the independent round-0 majority is typically reliable; we use it to establish where control reduces to voting and to locate any accuracy gains. A second tier of \emph{boundary probes} deliberately varies round-0 reliability and recoverability to characterize where control can help at all: a cross-domain extension to MATH-500 Level~5 and three further regimes (adversarial NLI, Winogrande, and a contested moral-judgment set), all introduced and analyzed together in Section~\ref{sec:boundary}. The probe tier is analyzed separately and is not pooled into the main-matrix results.

\paragraph{Shared round-0 baseline.}
All conditions share an identical round~0. We generate each agent's independent round-0 answer once per (task, seed) from a neutral, policy-agnostic prompt and reuse it across every condition; controllers act only from round~1 onward. The round-0 majority vote (R0-vote) is therefore the same controller-free starting point for every condition. The difference between a controller's final answer and R0-vote then measures what its round-1+ interaction (revision, reopening, synthesis) contributes, not how round~0 was framed.

All experiments use the same episode/round execution loop starting from this shared round~0, after which the policy may issue control actions from the core control space of Section~\ref{sec:method} (\texttt{explore}, \texttt{revise}, \texttt{accept}, \texttt{synthesize}, \texttt{justify}), with situational adding a \texttt{reconsider} extension; each policy uses a subset. The inclusive-leadership probe adds two further actions (\texttt{object}, \texttt{defend}) and is reported separately in Appendix~\ref{app:focused_social_core}. We log task identifiers, seeds, agent outputs by round, control actions, final answers, and token counts. Each (model, regime) combination is evaluated over three seeds (closed-ended, AlphaNLI, and broad Social Chemistry use 100-item sets; the mixed workload aggregates 150 items), and the MATH-500 extension uses five seeds. All three models are decoded at \texttt{temperature}=0.35; the remaining generation settings (token limits, round budgets, top-$p$ defaults) are detailed in Appendix~\ref{app:execution_settings}.

We also include a mixed-workload benchmark of 150 tasks: 50 closed-ended items, 50 AlphaNLI items, and 50 Social Chemistry items. Regime identity is not exposed to the controller, making this a direct test of adaptive leadership under heterogeneous deployment conditions.
\subsection{Evaluation and Analysis}
\label{sec:evaluation}
\label{sec:evaluation_axes}
We evaluate each policy along three axes: task performance, coordination cost, and behavioral mechanism. Task performance is measured with exact-match accuracy. Coordination cost is measured through interaction rounds and token cost. Behavioral mechanism is measured from logged trajectories through exploration rate, dissent preservation, majority lock-in, and recovery from bad round-0 consensus. Dissent preservation (whether a correct minority answer survives into the final synthesis) is reflected in the recovery rate and probed directly by the inclusive-leadership ablation (Appendix~\ref{app:focused_social_core}). The main tables foreground three highest-signal indicators, all per-run rates: \emph{majority lock-in}, the fraction of items whose final answer equals the independent round-0 majority; \emph{exploration}, the fraction of episodes in which the controller issues an \texttt{explore} action; and \emph{recovery}, the fraction of incorrect round-0 majorities the controller repairs ($P(\text{final correct}\mid\text{round-0 majority wrong})$).

\subsection{Predictions}
\label{sec:predictions}
Because each controller operationalizes a leadership theory (Section~\ref{sec:method}), that theory's own claim about when the style is needed transfers to a prediction about the controller, fixed in advance rather than read off our implementation. Because transactional leadership \citep{bass1985leadership} enforces an existing standard but offers no mechanism to find a better one, we predict it locks in the round-0 majority: efficient and matching the vote where that majority is reliable, brittle where it is wrong. Transformational \citep{bass1985leadership} and situational \citep{hersey1969management} leadership add, respectively, exploration around a shared goal and direction matched to the team's readiness to improve; both should break lock-in and recover incorrect majorities, but only where the majority is wrong, a better answer is reachable, and plain interaction has not already found it. Situational should gain most, since its readiness-based trigger reopens deliberation exactly when the team shows genuine disagreement. These predictions are inherited from theory, not tuned to the data: together they imply a largely null accuracy result punctuated by a single localized gain, with the controllers separating on behavioral signatures (lock-in, exploration, recovery) even where accuracy does not. Section~\ref{sec:theory_mapping} returns after the results to ground each measured region in the contingency construct it instantiates.

\section{Results}

\subsection{Behavioral Mechanism Analysis}
\label{sec:behavioral_mechanism}
As behavioral signatures are this work's primary scientific object, we present them first. Table~\ref{tab:behavior} summarizes the cross-model AlphaNLI readout for three high-signal indicators: explicit exploration, majority lock-in, and recovery from an incorrect round-0 consensus. Appendix~\ref{app:closed_ended_regime}, Appendix~\ref{app:alphanli_regime}, and Appendix~\ref{app:social_regime} report the corresponding regime-level follow-ups.

\begin{table*}[t]
\centering
\small
\resizebox{\textwidth}{!}{
\begin{tabular}{lcccccccccccc}
\toprule
\textbf{Condition} & \multicolumn{4}{c}{\textbf{\texttt{gpt-oss-120b}}} & \multicolumn{4}{c}{\textbf{\texttt{llama-4-scout}}} & \multicolumn{4}{c}{\textbf{\texttt{gemma-4-31B-it}}} \\
\cmidrule(lr){2-5} \cmidrule(lr){6-9} \cmidrule(lr){10-13}
 & Acc. & Explore & Lock-In & Recovery & Acc. & Explore & Lock-In & Recovery & Acc. & Explore & Lock-In & Recovery \\
\midrule
Flat                       & 0.920 & 0.000 & 0.977 & 0.130 & 0.873 & 0.000 & 0.960 & 0.175 & 0.950 & 0.000 & 0.980 & 0.286 \\
Theory-free control        & 0.903 & 0.000 & 0.963 & 0.087 & 0.880 & 0.000 & 0.953 & 0.225 & 0.920 & 0.000 & 0.990 & 0.000 \\
Transactional leadership   & 0.923 & 0.000 & \textbf{1.000} & 0.000 & 0.867 & 0.000 & \textbf{1.000} & 0.000 & 0.930 & 0.000 & \textbf{1.000} & 0.000 \\
\hspace{1em}\emph{accept-only ablation} & 0.923 & 0.000 & 1.000 & 0.000 & 0.867 & 0.000 & 1.000 & 0.000 & 0.930 & 0.000 & 1.000 & 0.000 \\
Transformational leadership& 0.903 & 0.100 & 0.963 & 0.087 & 0.873 & 0.153 & 0.953 & 0.200 & 0.927 & 0.063 & 0.990 & 0.048 \\
\hspace{1em}\emph{directives-only ablation} & 0.917 & 0.100 & 0.973 & 0.087 & 0.880 & 0.153 & 0.960 & 0.200 & 0.920 & 0.063 & 0.990 & 0.000 \\
Situational leadership     & 0.923 & \textbf{0.167} & 0.980 & 0.130 & 0.893 & \textbf{0.340} & 0.953 & 0.275 & 0.917 & \textbf{0.273} & 0.980 & 0.048 \\
\bottomrule
\end{tabular}
}
\caption{Cross-model behavioral metrics on AlphaNLI ($n=300$ per condition, 3 seeds). All three model families show the same ordering: transactional control has the highest majority lock-in (bold) with zero exploration, while situational shows the highest exploration (bold). The two indented rows are the Bass two-component ablations of Section~\ref{sec:bass_empirical}.}
\label{tab:behavior}
\end{table*}

\begin{figure}[t]
\centering
\includegraphics[width=0.95\columnwidth]{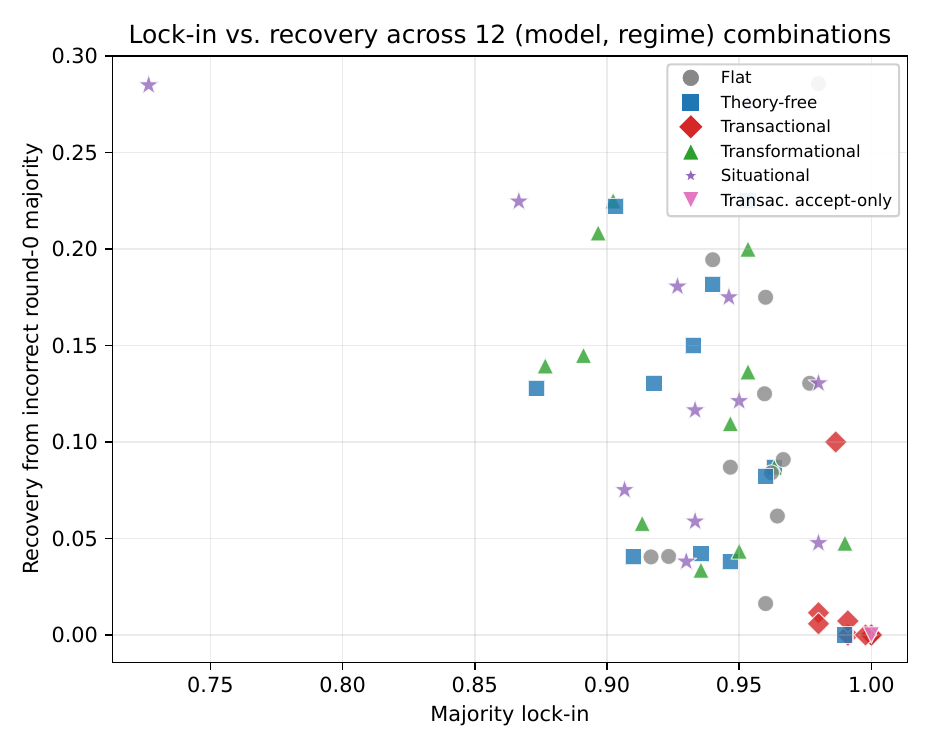}
\caption{Lock-in vs.\ recovery across the 12 (model, regime) combinations (3 model families $\times$ 4 regimes). Transactional control and its accept-only ablation cluster in the lower-right lock-in zone in every combination; the other controllers spread toward higher recovery and lower lock-in.}
\label{fig:lockin_recovery}
\end{figure}

The same ordering holds across all three model families, as Section~\ref{sec:predictions} predicted: transactional control locks in the round-0 majority at or near $1.000$ with essentially zero exploration and near-zero recovery, while the exploration-oriented controllers (situational, transformational) trade lock-in for higher exploration and recovery. Transactional's lock-in is the highest of any controller in every (model, regime) combination under a paired permutation test ($p<0.05$; most $p<0.001$). Figure~\ref{fig:lockin_recovery} shows this across all (policy, model, regime) points: transactional and its accept-only ablation sit in the lower-right lock-in zone everywhere, while the other controllers spread toward higher recovery and lower lock-in.

The reopening profile is strongest on the weakest model, \texttt{llama-4-scout}, where situational reaches the highest exploration ($0.34$ on AlphaNLI, $0.42$ on the social regime) and recovery (up to $0.29$ on social). It compresses on the stronger \texttt{gpt-oss-120b} and \texttt{gemma-4-31B-it}, whose independent round-0 reasoning is reliable enough (AlphaNLI round-0 vote $\geq 0.92$) that little incorrect consensus remains to recover from. The social regime is the clearest test of dissent preservation: on \texttt{llama-4-scout}, situational control reaches $0.513$ accuracy\footnote{This headline combination appears as $0.513$ in the main matrix, $0.517$ in some appendix probes (Appendices~\ref{app:mad_siturand}, \ref{app:focused_social_core}), and $0.527$ in the budget-6 and sensitivity runs (Appendices~\ref{app:round_budget}, \ref{app:situational_sensitivity}). The appendix probes are independent runs with their own generated-once shared round~0 whose main-condition accuracies differ from the main matrix by at most $\sim$2pp; we keep each figure as-run rather than reconcile them.} against transactional's $0.430$ and flat's $0.433$, with the lowest lock-in ($0.727$) and highest recovery ($0.285$), and transformational reaches $0.483$ (both gains significant over the shared round-0 vote after correction; Section~\ref{sec:per_controller_gain}). On the stronger models the social accuracies compress and no controller separates significantly, but the behavioral ordering persists.

\subsection{Performance and Cost Comparison}
\begin{figure*}[t]
\centering
\includegraphics[width=\textwidth]{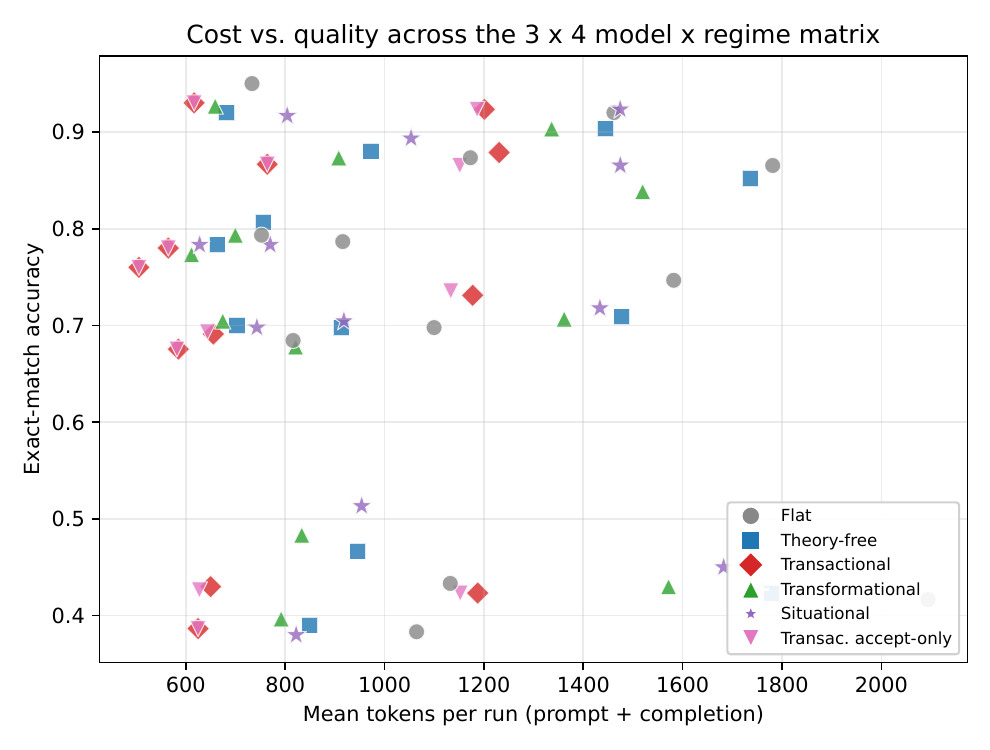}
\caption{Cost-quality Pareto across the $3 \times 4$ model-family $\times$ regime matrix. Each marker is one (policy, model, regime) combination; $x$ = mean prompt+completion tokens per run, $y$ = exact-match accuracy. The transactional accept-only ablation terminates at round~0 by construction and is the cheapest condition; situational takes the Pareto frontier on the \texttt{llama-4-scout} social regime.}
\label{fig:pareto}
\end{figure*}

The accuracy results complement the behavioral analysis. We read each controller against two baselines, weakest first. The \emph{round-0 vote} is the simplest possible aggregation of the shared independent draws; beating it shows that a controller's round-1+ interaction adds anything at all. \emph{Flat} is undirected multi-round deliberation; beating it is the stricter test, since it shows the controller's \emph{structure} adds value over plain interaction. We report both, and treat the gain over flat as the decisive one. Table~\ref{tab:main_results} reports the cross-model accuracy across the four regimes; Appendix~\ref{app:main_stop_rounds} reports stopping rounds, and Figure~\ref{fig:pareto} the matched cost-quality Pareto view (mean tokens per run vs.\ accuracy for every (model, regime, condition) combination), with numeric token costs in Appendix~\ref{app:tokens}.

\begin{table*}[t]
\centering
\small
\resizebox{\textwidth}{!}{
\begin{tabular}{lcccccccccccc}
\toprule
\textbf{Condition} & \multicolumn{4}{c}{\textbf{\texttt{gpt-oss-120b}}} & \multicolumn{4}{c}{\textbf{\texttt{llama-4-scout}}} & \multicolumn{4}{c}{\textbf{\texttt{gemma-4-31B-it}}} \\
\cmidrule(lr){2-5} \cmidrule(lr){6-9} \cmidrule(lr){10-13}
 & Closed & AlphaNLI & Social & Mixed & Closed & AlphaNLI & Social & Mixed & Closed & AlphaNLI & Social & Mixed \\
\midrule
Flat & 0.865 & 0.920 & 0.417 & 0.747 & 0.787 & 0.873 & 0.433 & 0.698 & 0.793 & 0.950 & 0.383 & 0.684 \\
Theory-free control & 0.852 & 0.903 & 0.423 & 0.709 & 0.807 & 0.880 & 0.467 & 0.698 & 0.783 & 0.920 & 0.390 & 0.700 \\
Transactional leadership & 0.879 & 0.923 & 0.423 & 0.731 & 0.780 & 0.867 & 0.430 & 0.691 & 0.760 & 0.930 & 0.387 & 0.676 \\
\hspace{1em}\emph{accept-only ablation} & 0.865 & 0.923 & 0.423 & 0.736 & 0.780 & 0.867 & 0.427 & 0.693 & 0.760 & 0.930 & 0.387 & 0.676 \\
Transformational leadership & 0.838 & 0.903 & 0.430 & 0.707 & 0.793 & 0.873 & \textbf{0.483} & 0.678 & 0.773 & 0.927 & 0.397 & 0.704 \\
\hspace{1em}\emph{directives-only ablation} & 0.855 & 0.917 & 0.420 & 0.704 & 0.787 & 0.880 & 0.453 & 0.682 & 0.783 & 0.920 & 0.373 & 0.684 \\
Situational leadership & 0.865 & 0.923 & 0.450 & 0.718 & 0.783 & 0.893 & \textbf{0.513} & 0.704 & 0.783 & 0.917 & 0.380 & 0.698 \\
\bottomrule
\end{tabular}
}
\caption{Cross-model exact-match accuracy across the four regimes ($n=300$ for the 100-item regimes, $n=450$ for the mixed workload; 3 seeds). Indented rows are the Bass two-component ablations of Section~\ref{sec:bass_empirical}. Bold marks the only two leadership conditions whose gain \emph{over the shared round-0 vote} is significant (\texttt{llama-4-scout} social; Table~\ref{tab:per_controller_gain}), not the within-column maximum, which is the usual convention elsewhere. Bootstrap 95\% CIs and pairwise paired permutation $p$-values in Appendix~\ref{app:statistics} and Table~\ref{tab:appendix_pairwise}.}
\label{tab:main_results}
\end{table*}

Table~\ref{tab:main_results} shows why a single universal ranking is the wrong summary. The accuracy spread across conditions is small on every (model, regime) combination except \texttt{llama-4-scout} social, where situational ($0.513$) and transformational ($0.483$) separate from the rest. Transactional usually stops in the fewest rounds (Appendix~\ref{app:main_stop_rounds}) but is never strongest in accuracy. On the easier regimes (closed-ended, AlphaNLI) the stronger models cluster near their round-0 ceiling and the controllers are statistically indistinguishable; on the mixed workload the nominal winner shifts by model (flat on \texttt{gpt-oss-120b}, situational on \texttt{llama-4-scout}, transformational on \texttt{gemma-4-31B-it}), with differences inside paired-test noise. We therefore read the behavioral signatures, which are stable across the matrix, as a more reliable basis of comparison than the absolute accuracy ordering. A peer-to-peer MAD baseline, run on all 12 combinations on the same shared round-0, never significantly beats a leadership controller: MAD tracks flat and the round-0 vote everywhere, and on the one combination with a real gain (\texttt{llama-4-scout} social) it loses to situational by $6.3$pp ($p=0.01$). A random-switch ablation further confirms that situational's social gain comes from its signal-driven trigger rather than its larger action space (Appendix~\ref{app:mad_siturand}).

The cost picture (Figure~\ref{fig:pareto}) reinforces this reading. Since accuracy is set mostly by regime and model rather than by controller, the conditions differ mainly in tokens spent: the transactional accept-only ablation, which stops at round~0, is the cheapest condition and transactional control is close behind, while the exploration-oriented controllers (situational, transformational) spend more for their extra rounds. On the eleven combinations where the controllers stay within accuracy noise, that extra spend buys nothing and the cheaper conditions sit on or near the cost-quality frontier; situational reaches the frontier only on \texttt{llama-4-scout} social, where its added cost is repaid by a real gain. The cost axis thus traces the same contingency boundary as the accuracy axis.

\subsection{Mixed-Workload Robustness}
The regime-separated results showed control helping in only one narrow regime, where the round-0 majority is unreliable, and reducing to voting elsewhere, with the controllers separating behaviorally rather than by accuracy. The next question is whether an adaptive controller can still recognize and help on that regime when the regimes are interleaved and their identity is hidden from it. To test that scenario, we construct a mixed 150-task benchmark that combines 50 closed-ended items, 50 AlphaNLI items, and 50 Social Chemistry items without exposing regime identity to the policy. The goal is to approximate a more realistic heterogeneous workload in which the controller must infer, from the task and the team's early interaction state alone, whether to converge quickly or preserve ambiguity.

The mixed-workload winner shifts by model and the aggregate gaps are small: flat leads on \texttt{gpt-oss-120b} ($0.747$), situational on \texttt{llama-4-scout} ($0.704$), and transformational on \texttt{gemma-4-31B-it} ($0.704$), each within paired-test noise of the runners-up. The behavioral picture, by contrast, is stable: transactional control keeps near-complete lock-in ($0.99$--$1.00$) and near-zero recovery on every model, while situational trades extra interventions for lower lock-in ($0.87$--$0.93$), higher recovery, and stronger performance on the social slice and the worst-source slice (Table~\ref{tab:appendix_mixed_source_breakdown}). As on the separated regimes, the exploration-oriented controllers help most where the round-0 majority is least reliable, and the accuracy ordering is less stable than the behavioral one.

\subsection{Per-controller gains over the round-0 vote}
\label{sec:per_controller_gain}

The absolute accuracy ordering shifts by regime and model, but a sharper question is which combinations show a leadership controller adding accuracy over the simplest possible aggregation. We compare each controller's final accuracy to the shared round-0 majority vote (R0-vote; Section~\ref{sec:experimental_setup}). Because round~0 is identical across conditions, a controller's gain over R0-vote measures exactly what its round-1+ interaction contributes over an independent draw plus a vote.

Table~\ref{tab:per_controller_gain} reports the gains. Transactional control reduces to the round-0 vote on all 12 (model, regime) combinations: its final accuracy is within $1.3$pp of R0-vote everywhere. On the main matrix its lock-in is thus accept-driven and its \texttt{revise} action dormant. Across the 36 leadership entries (three controllers $\times$ four regimes $\times$ three models), only two exceed a $5$pp gain that survives Benjamini--Hochberg correction, both on \texttt{llama-4-scout} (the lowest-capability model) on the social regime, the one combination where its round-0 majority is unreliable (round-0 vote accuracy $R_0 \approx 0.43$): situational ($+8.7$pp; bootstrap 95\% CI $[+3.3,+14.0]$; $q=0.05$) and transformational ($+5.7$pp; CI $[+2.3,+9.3]$; $q=0.05$). The stronger \texttt{gpt-oss-120b} and \texttt{gemma-4-31B-it} show no significant gain even on the social regime.

\begin{table}[t]
\centering
\small
\begin{tabular}{llcccc}
\toprule
\textbf{Model} & \textbf{Regime} & \textbf{R0-vote} & \textbf{Transac.} & \textbf{Transform.} & \textbf{Situa.} \\
\midrule
\texttt{gpt-oss-120b}  & AlphaNLI & 0.923 & $+0.000$ & $-0.020$ & $+0.000$ \\
\texttt{gpt-oss-120b}  & Closed   & 0.865 & $+0.013$ & $-0.027$ & $+0.000$ \\
\texttt{gpt-oss-120b}  & Social   & 0.423 & $+0.000$ & $+0.007$ & $+0.027$ \\
\texttt{gpt-oss-120b}  & Mixed    & 0.736 & $-0.004$ & $-0.029$ & $-0.018$ \\
\texttt{llama-4-scout} & AlphaNLI & 0.867 & $+0.000$ & $+0.007$ & $+0.027$ \\
\texttt{llama-4-scout} & Closed   & 0.780 & $+0.000$ & $+0.013$ & $+0.003$ \\
\texttt{llama-4-scout} & Social   & 0.427 & $+0.003$ & $\mathbf{+0.057}$\sigstar & $\mathbf{+0.087}$\sigstar \\
\texttt{llama-4-scout} & Mixed    & 0.693 & $-0.002$ & $-0.016$ & $+0.011$ \\
\texttt{gemma-4-31B-it}   & AlphaNLI & 0.930 & $+0.000$ & $-0.003$ & $-0.013$ \\
\texttt{gemma-4-31B-it}   & Closed   & 0.760 & $+0.000$ & $+0.013$ & $+0.023$ \\
\texttt{gemma-4-31B-it}   & Social   & 0.387 & $+0.000$ & $+0.010$ & $-0.007$ \\
\texttt{gemma-4-31B-it}   & Mixed    & 0.676 & $+0.000$ & $+0.029$ & $+0.022$ \\
\bottomrule
\end{tabular}
\caption{Accuracy gain of each leadership controller over the shared round-0 vote on the 12 (model, regime) combinations. The R0-vote column is the shared round-0 majority accuracy. \sigstar\ marks a gain $\geq 5$pp that survives Benjamini--Hochberg correction across all 36 leadership entries ($q<0.05$). Transactional gains nowhere (within $1.3$pp of R0-vote everywhere); both significant gains fall on the lowest-capability model on the social regime.}
\label{tab:per_controller_gain}
\end{table}

The behavioral signatures at the two gain combinations match the accuracy effect (Table~\ref{tab:gain_signatures}) and map onto the controllers' theoretical action sets. On \texttt{llama-4-scout} social, situational control breaks majority lock-in (0.73 vs.\ transactional's 0.98), explores in $0.42$ of episodes, and recovers $0.29$ of incorrect round-0 majorities, against transactional's near-zero recovery (0.01); transformational shows the same direction more mildly (lock-in 0.88, explore 0.29, recovery 0.14) via broadcast-goal exploration. Situational's active reopening and transformational's broadcast-goal direction are the levers that move accuracy beyond the round-0 vote; transactional, which only accepts, cannot.

\begin{table}[t]
\centering
\small
\begin{tabular}{lccc}
\toprule
\textbf{Condition (\texttt{llama-4-scout} social)} & \textbf{Lock-In} & \textbf{Explore} & \textbf{Recovery} \\
\midrule
Flat                & 0.923 & 0.000 & 0.041 \\
Theory-free control & 0.873 & 0.000 & 0.128 \\
Transactional       & 0.980 & 0.000 & 0.006 \\
Transformational    & 0.877 & 0.290 & 0.140 \\
Situational         & 0.727 & 0.420 & 0.285 \\
\bottomrule
\end{tabular}
\caption{Behavioral signatures on the \texttt{llama-4-scout} social regime, where the two controller-specific accuracy gains appear. Lock-In: $P(\text{final}=\text{round-0 majority})$. Explore: fraction of episodes with an explicit exploration action. Recovery: $P(\text{final correct}\mid\text{round-0 majority wrong})$.}
\label{tab:gain_signatures}
\end{table}

The methodology therefore separates two dimensions the controller label collapses: each controller has a reproducible behavioral signature (transactional locks in, situational actively reopens, transformational preserves dissent under a broadcast goal), but an \emph{accuracy} gain over the round-0 vote appears only where the round-0 majority is unreliable enough for reopening to pay off. Here that means the weakest model on the social regime ($R_0 \approx 0.43$). Where the round-0 majority is already reliable (closed-ended and AlphaNLI at $R_0 \geq 0.78$, and the stronger models throughout), every controller compresses to within a few points of R0-vote, and transactional coincides with it. That such gains are rare is what Section~\ref{sec:predictions} predicted; Section~\ref{sec:boundary} characterizes that boundary in detail, locating where a controller can beat plain interaction at all and why.

The entire positive accuracy evidence in this study comes down to this one (model, regime) combination. The paper's claims therefore rest on the mechanistic account: the recovery/breakage decomposition and its predictive test (Sections~\ref{sec:recovery_breakage} and~\ref{sec:predictive_analysis}), and the arbitrary-control baseline that separates the theory-derived rule from mere access to the action vocabulary (Section~\ref{sec:recovery_breakage}, Appendix~\ref{app:random_action}), both of which hold across the matrix and on hard math. The single positive combination is where the account says a gain should appear, but it remains a single observation (Section~\ref{sec:limitations}).

\section{Characterizing the Boundary: Recovery Advantage and a Cross-Domain Extension}
\label{sec:boundary}
Section~\ref{sec:per_controller_gain} found accuracy gains over the vote in only two of the 36 leadership entries, both on a single (model, regime) combination. This section characterizes that scarcity directly and sharpens it: against the stricter flat baseline, even that gain survives in only one combination. Using the boundary-probe tier (Section~\ref{sec:experimental_setup}), we decompose a controller's gain into recovery and breakage, define the recovery advantage that decides when control beats plain interaction, test it across four probe regimes (including the cross-domain MATH-500 extension), and read the resulting boundary against contingency theory.

\subsection{Why gains over plain interaction are rare: a recovery-advantage account}
\label{sec:recovery_breakage}
\begin{figure}[h]
\centering
\includegraphics[width=0.82\columnwidth]{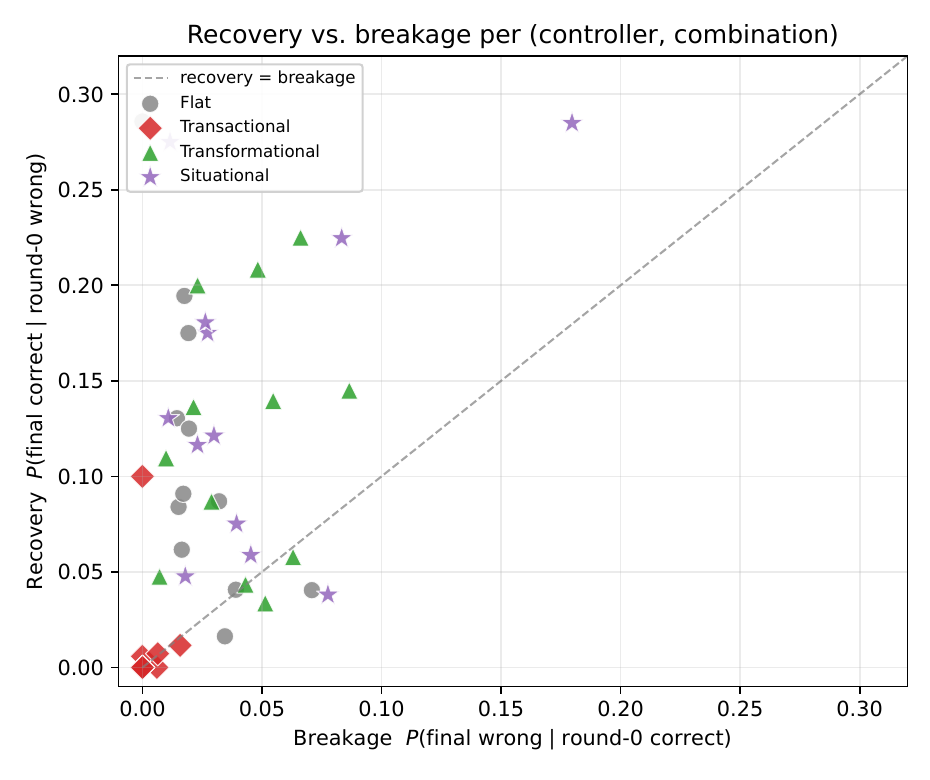}
\caption{Recovery vs.\ breakage for each (controller, combination). Transactional control sits at the origin: it never moves and thus equals the round-0 vote. Situational and transformational lie above the $y=x$ line (they recover more than they break); the single situational point below the line is \texttt{gemma-4-31B-it} social, where the model cannot repair its own errors.}
\label{fig:recovery_breakage}
\end{figure}

A controller's gain over the shared round-0 vote decomposes exactly into a benefit and a cost,
\[ \text{gain over the vote} = P(r_0\text{ wrong})\cdot\text{recovery} \;-\; P(r_0\text{ right})\cdot\text{breakage}, \]
where recovery $= P(\text{final correct}\mid\text{round-0 majority wrong})$ is the share of incorrect round-0 majorities the controller repairs, and breakage $= P(\text{final wrong}\mid\text{round-0 majority correct})$ is the share of correct majorities it drags to error. Each product weights a conditional rate by how often its case arises, and the two terms are thus the shares of all items the controller flips from wrong to right and from right to wrong; their difference is the controller's accuracy minus the vote's, which we report in percentage points. Pooled over the $1{,}194$ incorrect-round-0 and $2{,}853$ correct-round-0 items of the main matrix (Table~\ref{tab:recovery_breakage}, Figure~\ref{fig:recovery_breakage}), transactional control has recovery $0.007$ and breakage $0.002$: it essentially never moves, which is precisely why it coincides with the round-0 vote. Situational and transformational control do real work (recovery $0.140$ and $0.106$ at a $0.041$ breakage cost), but a net gain requires the benefit to outweigh the cost, which for situational happens only once $P(r_0\text{ wrong})$ exceeds about $0.23$.

\begin{table}[h]
\centering
\small
\begin{tabular}{lcc}
\toprule
\textbf{Condition} & \textbf{Recovery} & \textbf{Breakage} \\
\midrule
Flat & 0.075 & 0.021 \\
Theory-free control & 0.097 & 0.034 \\
Transactional & 0.007 & 0.002 \\
Transformational & 0.106 & 0.041 \\
Situational & 0.140 & 0.041 \\
\bottomrule
\end{tabular}
\caption{Recovery and breakage pooled over the 12 main-matrix combinations ($1{,}194$ incorrect-round-0 and $2{,}853$ correct-round-0 items). Transactional neither recovers nor breaks (it is the round-0 vote); situational recovers the most.}
\label{tab:recovery_breakage}
\end{table}

The two significant gain combinations are consistent with this account. They are exactly the combinations where the round-0 majority is unreliable enough \emph{and} the model can recover. The evidence is asymmetric. The account's null predictions are corroborated broadly (across every other combination, and out of sample in Section~\ref{sec:predictive_analysis}), while its positive region is observed once.

Beating the round-0 vote, however, is not the same as beating plain interaction. Since flat deliberation also recovers some incorrect round-0 majorities, the question that decides whether a \emph{controller} earns its structure is sharper: does it repair incorrect majorities that flat leaves unrepaired? We therefore read the \emph{recovery advantage} $\Delta_{\text{rec}} = \text{recovery(situational)} - \text{recovery(flat)}$. A controller's accuracy gain over flat appears only when $\Delta_{\text{rec}}$ is large and positive, which, as the boundary probes below show, is rare.

\paragraph{Boundary probes.} To locate where an accuracy gain over flat is possible, we probe four further regimes chosen to vary the two governing quantities independently (full per-model results in Appendix~\ref{app:boundary_probes}). Two have objective gold that keeps the 3-agent round-0 majority \emph{reliable} even when individual agents are fooled: adversarial natural-language inference (ANLI-R3 \citep{nie2020adversarial}) and Winogrande \citep{sakaguchi2021winogrande}. One has subjective, contested gold that makes the round-0 majority \emph{unreliable but unrecoverable}: a moral-judgment regime of 99 balanced, genuinely contested \texttt{r/AmITheAsshole} verdicts from Scruples \citep{lourie2021scruples} (median community agreement $0.56$), where predicting the crowd's plurality is not derivable from the text by reasoning. The per-item agreement distribution supports reading this regime as unrecoverable. Across the 99 items the plurality share of community votes has mean $0.55$ and maximum $0.69$. Even an idealized judge drawn from the same community would therefore match the gold plurality at an expected rate of only about $0.55$, a ceiling close to the observed round-0 vote accuracies. Residual errors here are dominated by genuine disagreement in the gold signal, not by reasoning failures that deliberation could repair. The fourth is MATH-500 Level~5 (Section~\ref{sec:math500}), whose round-0 is unreliable and \emph{recoverable but already exploited by flat}. Table~\ref{tab:boundary} lines these up against the one regime where a controller wins. The recovery advantage $\Delta_{\text{rec}}$ orders the accuracy gain across all of them: it is near zero wherever the round-0 majority is reliable (ANLI, Winogrande: little incorrect consensus to repair), near zero wherever the task is unrecoverable (Scruples, $\Delta_{\text{rec}}\le 0.01$ on every model; \texttt{gemma-4-31B-it} social), and negative wherever flat already recovers (MATH, $\Delta_{\text{rec}}<0$); it is large ($+0.24$) only on \texttt{llama-4-scout} social, the single combination where the model repairs incorrect majorities that flat cannot.

\begin{table}[h]
\centering
\small
\begin{tabular}{llccccc}
\toprule
\textbf{Regime} & \textbf{Model} & $P(r_0\text{ wrong})$ & Recov.\ flat & Recov.\ situ. & $\Delta_{\text{rec}}$ & $\Delta$acc \\
\midrule
Social & \texttt{llama-4-scout} & 0.57 & 0.04 & 0.28 & $+0.24$ & $+0.080$\sigstar \\
Social & \texttt{gemma-4-31B-it} & 0.61 & 0.02 & 0.04 & $+0.02$ & $-0.003$ \\
Scruples & \texttt{llama-4-scout} & 0.61 & 0.02 & 0.02 & $+0.00$ & $-0.010$ \\
Scruples & \texttt{gpt-oss-120b} & 0.55 & 0.03 & 0.04 & $+0.01$ & $+0.000$ \\
ANLI-R3 & \texttt{llama-4-scout} & 0.27 & 0.21 & 0.22 & $+0.01$ & $-0.063$ \\
Winogrande & \texttt{gemma-4-31B-it} & 0.10 & 0.20 & 0.17 & $-0.03$ & $-0.020$ \\
MATH-500 & \texttt{llama-4-scout} & 0.40 & 0.41 & 0.28 & $-0.13$ & $-0.056$ \\
MATH-500 & \texttt{gemma-4-31B-it} & 0.22 & 0.76 & 0.39 & $-0.37$ & $-0.088$ \\
\bottomrule
\end{tabular}
\caption{Boundary probes: a controller's accuracy gain over flat ($\Delta$acc $=$ situational $-$ flat, paired permutation) tracks its \emph{recovery advantage} $\Delta_{\text{rec}}=$ recovery(situational)$-$recovery(flat), not round-0 unreliability $P(r_0\text{ wrong})$ alone. A leadership-specific gain requires all three of an unreliable round-0, a recoverable task, and a recovery flat does not already realize, met only on \texttt{llama-4-scout} social. This $\Delta$acc is the gain over \emph{flat} ($+0.080$ on that combination); the $+8.7$pp quoted elsewhere is the same combination's gain over the round-0 \emph{vote} (Table~\ref{tab:per_controller_gain}), a weaker baseline. \sigstar: significant after correction.}
\label{tab:boundary}
\end{table}

The recovery term depends on model capability, which is why the win is narrow even among unreliable-round-0 regimes. On the social regime \texttt{llama-4-scout} recovers $0.28$ of its incorrect majorities against $0.18$ breakage (net $+8$pp over the vote), whereas \texttt{gemma-4-31B-it} recovers only $0.04$ against $0.08$ breakage (net $-1$pp): it cannot repair its own social errors, and reopening does not pay off even though its round-0 is equally unreliable. Appendix~\ref{app:case_studies} gives representative case-study transcripts of a successful recovery, a failed recovery, and a breakage from this combination, the qualitative face of these pooled rates. Gains over plain interaction are therefore governed by quantities the controller label hides (how often the round-0 majority is wrong, whether the task is recoverable at all, and whether the controller recovers where flat does not), and the methodology measures each of them. Their intersection is small, which is exactly why accuracy gains are rare; the recovery signature, not the controller label, is what locates it.

The recovery comes from the controllers' theory-derived decision rules. We verify this with an \emph{arbitrary-control} baseline that uses the same control actions but assigns them uniformly at random each round, with no leadership-derived structure (Table~\ref{tab:random_action_main}; run details in Appendix~\ref{app:random_action}). This is the control that separates ``the action vocabulary did it'' from ``the theory-derived rule did it.'' On a matched run that adds it alongside the leadership controllers, the arbitrary controller recovers $0.009$ of incorrect round-0 majorities: indistinguishable from the round-0 vote ($0.010$) and far below situational's $0.133$ and transformational's $0.089$ on the same items, with situational's recovery exceeding it in all 9 combinations. It even falls below flat ($0.061$), because random actions churn the team without coherently reopening a wrong majority. The same action set used arbitrarily has no effect; the theory-derived rule for \emph{when} to reopen and \emph{what} to broadcast is what produces recovery.

\begin{table}[h]
\centering
\small
\begin{tabular}{lcc}
\toprule
\textbf{Condition} & \textbf{Recovery} & \textbf{Breakage} \\
\midrule
Flat & 0.061 & 0.029 \\
Transactional & 0.010 & 0.003 \\
Transformational & 0.089 & 0.037 \\
Situational & 0.133 & 0.043 \\
Arbitrary control (random action) & 0.009 & 0.003 \\
\bottomrule
\end{tabular}
\caption{Arbitrary-control baseline: recovery and breakage pooled over the AlphaNLI, social, and mixed regimes on the three models ($1{,}021$ incorrect-round-0 items). Despite the same action set, the arbitrary controller recovers no better than the round-0 vote.}
\label{tab:random_action_main}
\end{table}

A second baseline brackets the question from the opposite side. Instead of the action vocabulary without a rule, we add an \emph{unconstrained coordinator}, a plain LLM that coordinates the team with free-form messages and no leadership principles, no vocabulary, and no threshold rule (Appendix~\ref{app:llm_coordinator}). Started from the identical shared round~0 with the same final aggregation, it reduces to the round-0 vote where the majority is reliable, is significantly harmful on all four \texttt{gemma-4-31B-it} regimes ($-5.7$ to $-8.3$pp against the vote, all $q<0.01$; breakage with near-zero recovery), and on the readiness-gap combination captures little of the available gain, staying $6.0$pp below situational control (paired $p=0.016$). Generic coordination is not what the theory-derived rule contributes.

These baselines also address a natural objection: that situational control might be no more than a generic ``reopen on disagreement'' heuristic dressed in leadership terms. A signal-matched reopening rule is not a \emph{non}-leadership control: state-contingent intervention \emph{is} the operational content of situational leadership \citep{hersey1969management}. We bracket it from two sides. Because a matched-rate but signal-\emph{blind} reopening rule (random-switch, Appendix~\ref{app:mad_siturand}) underperforms situational, the gain comes from reopening at the right times, not from reopening alone; and the sensitivity analysis (Appendix~\ref{app:situational_sensitivity}) localizes the gain to the 2--1-split trigger. We make no claim that the differentiated directives beyond that reopening decision are themselves load-bearing; the theory's contribution is the decision rule for \emph{when} to reopen.

The cross-domain extension below pushes round-0 reliability to its extremes, where this recovery advantage turns negative and pins down what a leadership-specific gain requires beyond an unreliable round-0.

\subsection{A formal predictive test of the recovery-advantage account}
\label{sec:predictive_analysis}
\begin{figure}[t]
\centering
\includegraphics[width=0.78\columnwidth]{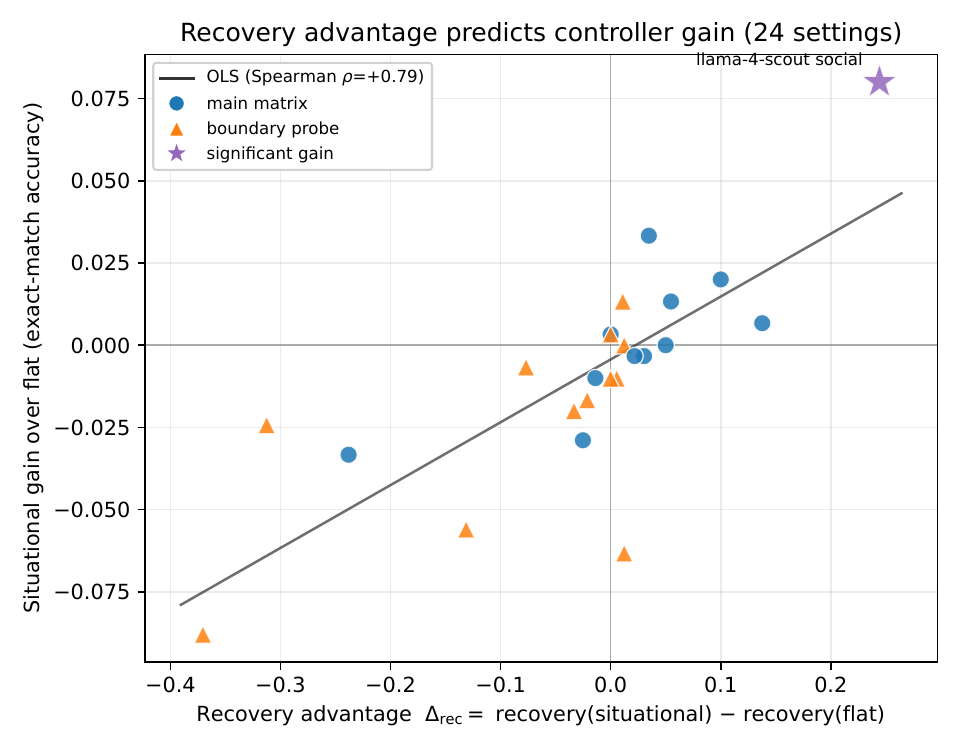}
\caption{The recovery advantage $\Delta_{\text{rec}}$ predicts the situational gain over flat across all 24 evaluated settings (Spearman $\rho=+0.789$, $p<0.001$); the single positive setting (\texttt{llama-4-scout} social) sits at the largest $\Delta_{\text{rec}}$.}
\label{fig:predictive}
\end{figure}

The qualitative reading above says that the recovery advantage $\Delta_{\text{rec}}$, not round-0 unreliability alone, orders where a controller beats plain interaction. We test this quantitatively across all 24 evaluated settings (the 12 main-matrix combinations plus the 12 boundary-probe combinations), with the situational gain over flat as the outcome. Figure~\ref{fig:predictive} shows that relationship. The released per-setting table carries every value.

The recovery advantage $\Delta_{\text{rec}}$ predicts the gain over flat with Spearman $\rho=+0.789$ (permutation $p<0.001$; bootstrap 95\% CI $[+0.51,+0.93]$); an OLS fit gives $+1.9$pp of gain per $+0.10$ of recovery advantage. By contrast, round-0 unreliability alone, the account's necessity condition, does not predict the gain ($\rho=+0.316$, $p=0.13$), and recoverability alone does not either ($\rho=-0.176$, $p=0.40$); flat recovery alone correlates \emph{negatively} ($\rho=-0.656$, $p=0.001$), since wherever undirected interaction already recovers there is nothing left for a controller to add. This is the formal version of the necessity-not-sufficiency claim: each governing quantity matters only through the composite $\Delta_{\text{rec}}$.

Because the boundary probes were chosen to span the governing quantities, pooling them with the main matrix could overstate the relationship. Restricted to the 12 main-matrix combinations, which were fixed before the account existed and were not selected on any predictor, the relationship is \emph{strongest} ($\rho=+0.858$, $p=0.001$); on the probes alone it is $+0.508$ ($p=0.10$). The relationship is therefore not an artifact of diagnostic probe selection.

Under leave-one-model-out and leave-one-task-out validation (fitting the gain on the remaining settings and predicting the held-out one), the null settings are predicted tightly: mean absolute error $1.5$pp over the 23 null settings for both schemes. The account therefore predicts \emph{where control does not help} out of sample. The single positive setting, held out, is under-predicted ($+2.8$ to $+3.9$pp against the observed $+8.0$pp) but remains top-ranked and positive, even when its entire model family is excluded from the fit, in which case the training settings contain no significant positive at all. This asymmetry is why we present the readiness gap as an explanatory hypothesis.

\subsection{Cross-domain extension: hard math (necessity is not sufficiency)}
\label{sec:math500}

The 12 main-matrix combinations span regimes in which per-agent round-0 reasoning is typically reliable. We now apply the same behavioral-signature and per-action-ablation methodology to MATH-500 Level~5 \citep{hendrycks2021measuring,lightman2024lets}, a competition-math benchmark (simple-numeric-answer subset, 50 items) on which round-0 reliability varies sharply across our three models. This is the regime in which the \texttt{revise} action of transactional control (Section~\ref{sec:bass_two_pillar}), dormant on the main matrix, actually fires; Section~\ref{sec:bass_empirical} uses this to decompose transactional control per action.

\subsubsection{Setup}
We evaluate the three models with 3-agent teams under the five main conditions plus the \texttt{transactional\_accept\_only} ablation (Section~\ref{sec:bass_empirical}), five seeds per condition ($n=250$). The cross-round majority extractor from Section~\ref{sec:cross_round_extraction} is the default finalize step; \texttt{gpt-oss-120b}, a reasoning model, uses a larger generation budget so that its round-0 reasoning is not truncated before committing a final answer (Appendix~\ref{app:execution_settings}).

\subsubsection{Behavioral signatures and accuracy on hard math}

\begin{table*}[t]
\centering
\small
\resizebox{\textwidth}{!}{
\begin{tabular}{lcccccccccccc}
\toprule
\textbf{Condition} & \multicolumn{4}{c}{\textbf{\texttt{gpt-oss-120b}}} & \multicolumn{4}{c}{\textbf{\texttt{llama-4-scout}}} & \multicolumn{4}{c}{\textbf{\texttt{gemma-4-31B-it}}} \\
\cmidrule(lr){2-5} \cmidrule(lr){6-9} \cmidrule(lr){10-13}
 & Acc. & Lock-In & Recov. & Stop@R & Acc. & Lock-In & Recov. & Stop@R & Acc. & Lock-In & Recov. & Stop@R \\
\midrule
Flat & 0.960 & 0.947 & 0.625 & 1.24 & 0.736 & 0.632 & 0.414 & 1.76 & 0.944 & 0.792 & 0.759 & 1.58 \\
Theory-free control & 0.944 & 0.943 & 0.500 & 1.32 & 0.700 & 0.668 & 0.283 & 2.03 & 0.912 & 0.792 & 0.611 & 1.72 \\
Transactional leadership & 0.936 & 0.951 & 0.312 & 1.17 & 0.720 & 0.732 & 0.323 & 1.46 & 0.900 & 0.836 & 0.556 & 1.26 \\
Transformational leadership & 0.936 & 0.947 & 0.312 & 1.20 & 0.664 & 0.664 & 0.212 & 1.56 & 0.864 & 0.784 & 0.426 & 1.42 \\
Situational leadership & 0.936 & 0.955 & 0.312 & 1.31 & 0.680 & 0.648 & 0.283 & 1.87 & 0.856 & 0.776 & 0.389 & 1.64 \\
\bottomrule
\end{tabular}
}
\caption{MATH-500 Level~5: accuracy and behavioral signatures, $n=250$ per condition (5 seeds). Accuracy is full-sample with the cross-round majority extractor (Section~\ref{sec:cross_round_extraction}); Lock-In and Recovery are computed over runs with a parseable round-0 majority; Stop@R is the mean stopping round. The \texttt{transactional\_accept\_only} ablation is reported in Table~\ref{tab:math500_accept_only}.}
\label{tab:math500_main}
\end{table*}

Table~\ref{tab:math500_main} reports per-condition accuracy and behavioral signatures across the three models. The three models span a wide range of round-0 reliability: the round-0 majority is incorrect in only $\sim 0.06$ of \texttt{gpt-oss-120b} runs, $\sim 0.22$ on \texttt{gemma-4-31B-it}, and $\sim 0.40$ on \texttt{llama-4-scout}. Where round-0 is reliable (\texttt{gpt-oss-120b}, round-0 vote $0.94$), all controllers cluster near the flat ceiling ($0.94$--$0.96$) and deliberation adds little. Where it is unreliable (\texttt{llama-4-scout}, round-0 vote $0.60$), multi-round deliberation recovers many incorrect round-0 majorities, lifting flat to $0.74$ over the round-0 vote.

The behavioral ordering matches the main matrix: transactional control, and especially its accept-only ablation (Section~\ref{sec:bass_empirical}), produces the highest lock-in and the lowest recovery, while flat and the exploration-oriented controllers recover more. Recovery itself scales with capability: every controller on \texttt{gpt-oss-120b} recovers from an incorrect round-0 majority more often than any controller on \texttt{llama-4-scout}. On \texttt{gpt-oss-120b} the leadership controllers nonetheless recover less than flat ($0.31$ vs.\ flat's $0.63$): they close on the 2-of-3 majority and forgo the extra rounds flat spends repairing the rare incorrect one, but with only $\sim 0.06$ of round-0 majorities wrong this costs almost no accuracy. The decisive comparison is the \texttt{revise}-action ablation, which isolates what the management-by-exception action contributes in each reliability regime (Section~\ref{sec:bass_hardmath}).

This MATH evidence sharpens the round-0-reliability axis into a \emph{necessity-not-sufficiency} statement. Where round-0 is unreliable, interaction adds value, but on hard math that value accrues to \emph{flat} interaction, not to the leadership controllers: flat is the most accurate condition on all three models, and every leadership controller trails it (situational by $2.4$/$5.6$/$8.8$pp on \texttt{gpt-oss-120b}/\texttt{llama-4-scout}/\texttt{gemma-4-31B-it}; transactional, the lock-in controller, is the strongest or tied-strongest leadership condition on every model). This is the opposite of \texttt{llama-4-scout} social, where situational beats flat by $8$pp (Table~\ref{tab:main_results}). The contrast pins down what a leadership-specific advantage requires beyond an unreliable round-0: a failure mode of premature consensus on genuinely ambiguous items, where structured reopening outperforms undirected interaction. On hard math the disagreement is largely self-correcting: a wrong numeric answer tends to expose its own error under scrutiny. Plain flat interaction thus already recovers incorrect round-0 majorities and leaves a controller little to add. Round-0 unreliability is therefore necessary but not sufficient for a leadership-specific gain; the gain also requires that undirected interaction cannot already exploit the unreliability on its own. We return to this boundary in the limitations (Section~\ref{sec:limitations}).

\subsection{Mapping the boundary to leadership contingency theory}
\label{sec:theory_mapping}
The boundary assembled across Sections~\ref{sec:recovery_breakage}--\ref{sec:math500} is not only an empirical regularity; it lines up with the contingency view of leadership we adopted as a null hypothesis. Team science holds that a leader adds value only by supplying a function the team needs and is not otherwise getting \citep{hackman1986leading}, and that several conditions make leadership unnecessary or impossible \citep{kerr1978substitutes}. Each region of our measured boundary corresponds to one of those conditions (Table~\ref{tab:theory_map}).

\begin{table}[h]
\centering
\small
\setlength{\tabcolsep}{4pt}
\begin{tabular}{p{0.26\columnwidth}p{0.14\columnwidth}p{0.27\columnwidth}p{0.18\columnwidth}}
\toprule
\textbf{Measured region} & \textbf{Signature} & \textbf{Contingency construct} & \textbf{Observe} \\
\midrule
Round-0 reliable & $P(r_0\text{ wrong})$ low & leadership substitutes \citep{kerr1978substitutes} & control $=$ vote \\
Unreliable, unrecoverable & recovery $\approx 0$ & no latent ability to organize \citep{hackman1986leading} & control $=$ vote \\
Unreliable, flat recovers & $\Delta_{\text{rec}}<0$ & path-goal redundancy \citep{house1971path} & control $\leq$ flat \\
Unreliable, recoverable, flat insufficient & $\Delta_{\text{rec}}\gg 0$ & readiness gap \citep{hersey1969management} & situational $>$ flat ($+8$pp) \\
\bottomrule
\end{tabular}
\caption{The measured boundary (Section~\ref{sec:recovery_breakage}, Table~\ref{tab:boundary}) read against leadership contingency theory. Each region instantiates a named construct, and the observed outcomes are consistent with the predictions of Section~\ref{sec:predictions}; a controller adds value only in the readiness-gap region, observed in a single combination.}
\label{tab:theory_map}
\end{table}

A reliable round-0 is the analogue of \emph{leadership substitutes}: the agents already solve the task and control reduces to the vote; an unrecoverable task (Scruples; \texttt{gemma-4-31B-it} social) offers no latent ability for a controller to organize; and where undirected interaction already recovers (hard math), directive control is \emph{redundant} in the path-goal sense. A controller earns its keep only in the remaining \emph{readiness gap} (an unreliable round-0 that the team can repair but flat does not), the one region (\texttt{llama-4-scout} social) where situational control gains.

All four regions of this boundary follow from contingency theory and are consistent with the predictions of Section~\ref{sec:predictions}. The reliability conditions motivated the shared-round-0 design, and the boundary probes bear out the recoverability conditions, contrasting the unrecoverable Scruples probe with recoverable math. The measurement vocabulary is a \emph{computational testbed} whose boundary is consistent with leadership contingency theory, not an adjudication between theories. The boundary itself carries the same qualification. Its null regions are corroborated broadly, including out of sample (Section~\ref{sec:predictive_analysis}); its readiness-gap region is observed in one combination on one low-capability model. We therefore advance the readiness gap as an explanatory hypothesis, with prospective validation on independently selected settings as the natural next test (Section~\ref{sec:limitations}).

\section{Bass Two-Component Empirical Decomposition}
\label{sec:bass_empirical}

Section~\ref{sec:bass_two_pillar} introduced two component-level ablations: \texttt{transactional\_accept\_only}, which removes the management-by-exception action from transactional control while keeping the contingent-reward action, and \texttt{transformational\_directives\_only}, which removes the broadcast-goal component from transformational control while keeping the differentiated-directive component. Each ablation isolates one component of a Bass classical leadership style at the action level, leaving the other component active. We report the transactional decomposition across the 12 main-matrix combinations and on MATH-500 Level~5 (Section~\ref{sec:math500}), together with the transformational two-component ablation across the four regimes as a design-symmetric companion.

\subsection{Main matrix: the management-by-exception action is structurally dormant}
\label{sec:bass_paper12}

On the main matrix the \texttt{revise} (management-by-exception) action is structurally dormant. Because transactional control stops at round~0 in nearly every episode (its mean stopping round is at most $1.03$ on all 12 combinations), the \texttt{revise} action is almost never available to alter the outcome. Two consequences follow. First, the transactional lock-in signature is carried entirely by the \texttt{accept} (contingent-reward) action: accept-only and full transactional converge in the same round~0 and accept-only inherits the same near-ceiling lock-in (Table~\ref{tab:behavior}). The signature is therefore \emph{accept-driven, not revise-driven}: a strictly stronger claim than ``revise contributes little,'' since revise is dormant by construction.

Second, with revise dormant the accept-only ablation coincides with the shared round-0 vote on all 12 combinations: the accept-only rows of Table~\ref{tab:main_results} match the R0-vote column of Table~\ref{tab:per_controller_gain} exactly. This confirms the controller-level reducibility result of Section~\ref{sec:per_controller_gain} at the component level: where its second action never fires, transactional control adds nothing over voting the shared round~0.

\subsection{Hard math: the revise action helps where round-0 is unreliable}
\label{sec:bass_hardmath}

On MATH-500, where round-0 disagreement is common, the \texttt{revise} action fires and its effect becomes measurable. Table~\ref{tab:math500_accept_only} contrasts full transactional with its accept-only ablation across the three models. Removing the revise action lowers accuracy on the two models whose round-0 majority is unreliable, \texttt{llama-4-scout} ($-8.0$pp, $p<0.01$) and \texttt{gemma-4-31B-it} ($-8.0$pp, $p<0.01$), because the revise step is what recovers incorrect round-0 majorities through later rounds. On \texttt{gpt-oss-120b}, whose round-0 majority is already reliable (round-0 vote $0.94$), removing revise has no effect: accept-only ($0.928$), full transactional ($0.936$), and the round-0 vote ($0.935$) coincide, all just below flat ($0.960$).

\begin{table}[h]
\centering
\small
\begin{tabular}{lcccccc}
\toprule
\textbf{Model} & R0-vote & Flat & Transac. & Accept-only & $\Delta_{\text{revise}}$ & $p$ \\
\midrule
\texttt{gpt-oss-120b}  & 0.935 & 0.960 & 0.936 & 0.928 & $-0.008$ & $0.68$ \\
\texttt{llama-4-scout} & 0.604 & 0.736 & 0.720 & 0.640 & $-0.080$ & ${<}0.01$ \\
\texttt{gemma-4-31B-it}   & 0.784 & 0.944 & 0.900 & 0.820 & $-0.080$ & ${<}0.01$ \\
\bottomrule
\end{tabular}
\caption{MATH-500 Level~5: effect of removing the \texttt{revise} action, $n=250$ per condition. $\Delta_{\text{revise}}=$ accept-only $-$ full transactional (paired permutation). Removing revise hurts the two models with unreliable round-0 majorities and has no effect on \texttt{gpt-oss-120b}, whose round-0 is already reliable; it never helps.}
\label{tab:math500_accept_only}
\end{table}

The revise action thus helps exactly where the round-0 majority is unreliable and has no effect where it is reliable: the same round-0-reliability threshold that makes the controller reducible to the round-0 vote on the main matrix. Removing revise never improves accuracy at any model scale: there is no direction reversal in which convergence pressure becomes a net asset on a stronger model. Where round-0 is reliable, the controller simply has nothing to add; where it is unreliable, the revise action does real work.

\subsection{Complementary decomposition: transformational two-component}
\label{sec:bass_transformational}

The same design applies to the transformational controller, decomposed into the broadcast-goal preamble and the differentiated-directive per-agent loop (Section~\ref{sec:bass_two_pillar}); the directives-only ablation rows appear in Table~\ref{tab:main_results}. The broadcast-goal component contributes on the ambiguity-heavy regimes and is null elsewhere: full transformational exceeds directives-only by $+3.0$pp on \texttt{llama-4-scout} social, $+2.4$pp on \texttt{gemma-4-31B-it} social, and $+2.0$pp on \texttt{gemma-4-31B-it} mixed, while on closed-ended QA and AlphaNLI the two coincide within $\pm 2$pp.

The two decompositions are regime-dependent at different mechanistic layers. The transactional decomposition is regime-dependent in \emph{whether the second action fires at all} (revise dormant on the main matrix, active on unreliable-round-0 math); the transformational decomposition is regime-dependent in \emph{which component carries the regime} (the broadcast goal on the ambiguity-heavy regimes, the differentiated directives elsewhere). Both land in the same Bass two-component frame.

\subsection{Synthesis}
\label{sec:bass_synthesis}

Combining the decompositions, the management-by-exception (\texttt{revise}) action is dormant where the round-0 majority is reliable (the main matrix, and \texttt{gpt-oss-120b} on hard math) and decisive where it is unreliable (\texttt{llama-4-scout} and \texttt{gemma-4-31B-it} on hard math), tracking the round-0-reliability axis of Section~\ref{sec:per_controller_gain}. The transformational broadcast-goal component shows the same regime dependence, contributing on the social and mixed regimes and null elsewhere.

What a per-action ablation reveals therefore depends on whether the component fires. Where it does, the methodology yields a mechanism-level finding (the revise action's reliability-gated value); where it does not, it yields a reducibility result (transactional equals the round-0 vote) that bounds how strongly the controller-level signature can be claimed. The two outcomes are the same measurement applied in opposite reliability regimes.

\section{Discussion}

\paragraph{Contingency, confirmed and measured.}
Our results support the contingency view we began with: there is no universal best controller, and process-level control adds value only in the restricted readiness-gap regime that the measurement vocabulary identifies and team-science theory predicts (Sections~\ref{sec:recovery_breakage}--\ref{sec:theory_mapping}). Across the three model families, absolute accuracy rankings shift more than the within-controller behavioral signatures. The stable scientific object is therefore the interaction signature a controller induces, not a single best policy.

\paragraph{Three patterns of regime fit.}
\emph{Transactional control} is the lock-in mechanism: its majority lock-in is the highest of any controller in every (model, regime) combination ($p<0.05$, most $p<0.001$; Appendix~\ref{app:statistics}), which makes it efficient for disciplined convergence and brittle under genuine ambiguity. The two are faces of the same accept-driven mechanism and the reason it coincides with the round-0 vote wherever that vote is already good. \emph{Situational control} is the clearest ambiguity-oriented mechanism: it produces the study's only significant accuracy gain ($+8.7$pp over the shared round-0 vote on \texttt{llama-4-scout} social, $q=0.05$) by breaking lock-in and recovering from incorrect round-0 majorities. \emph{Transformational control} shows the same direction more mildly ($+5.7$pp on the same combination) via broadcast-goal exploration. On the stronger \texttt{gpt-oss-120b} and \texttt{gemma-4-31B-it}, whose round-0 majorities are already reliable, all three controllers converge toward the round-0 vote and separate only in their behavioral signatures. In deployment terms, process-level control earns its token overhead only where the independent round-0 majority is unreliable yet recoverable; elsewhere plain voting or flat interaction already sits on the cost-quality frontier.

\paragraph{Component axis: Bass two-component decomposition is regime-dependent.}
The third axis is internal to each controller. The Bass two-component ablations show that the transactional controller's main-matrix signature is carried entirely by its \emph{accept} (contingent-reward) action: the \texttt{revise} action is dormant (transactional stops at round~0 in nearly every episode), and removing it leaves both the lock-in signature and the round-0-vote accuracy intact. On MATH-500 Level~5, where round-0 majorities are unreliable for the weaker models, the \texttt{revise} action does fire and removing it lowers accuracy by $\sim 8$pp ($p<0.01$); on the model whose round-0 is already reliable it has no effect, and it never helps. Transformational control shows the complementary pattern: its broadcast-goal component contributes on the ambiguity-heavy regimes and is null elsewhere. The component axis therefore obeys the same round-0-reliability rule as the controller axis: a second action earns its keep only where the first leaves an unreliable majority to repair.

\paragraph{Process-level scope, knowledge-level complementarity.}
The current study is intentionally restricted to \emph{process-level} coordination on symmetric-input shared-information teams of three agents. \emph{Knowledge-level} coordination patterns from the MAS literature (multi-agent debate, role specialization, pipelined decomposition, self-refinement \citep{du2024improving,liang2024encouraging,chan2024chateval,hong2024metagpt,wu2023autogen,li2023camel,chen2023agentverse,madaan2023selfrefine,shinn2023reflexion}) are deliberately not the manipulated variable here. They remain complementary and can compose with the process-level vocabulary we develop. Asymmetric topologies (subtask decomposition, sequential pipelines, specialist-generalist splits) and agentic tasks that require tool use or multi-step execution introduce additional design degrees of freedom that would each plausibly explain any observed signature difference. Characterizing controller mechanism in the cleanest possible setting is a prerequisite for interpreting controller effects in those more confounded deployments.

\paragraph{Leadership as a starting point.}
Leadership theory is one entry into a broader research program that imports theory from team science into multi-agent LLM coordination. Conflict regulation, expertise recognition, trust calibration, and adaptive teaming each carry their own theoretical structure that could be mapped onto explicit control actions and probed with the same behavioral-signature and per-action ablation vocabulary developed here. Whether each admits such a faithful mapping in LLM agents, rather than only in human teams, is itself an empirical question that each case study must establish, not assume. We see leadership as a starting point rather than a destination.

\paragraph{Limitations.}
\label{sec:limitations}
The paper has several limitations. The evaluation is built around regime-focused subsets and reference pools rather than full benchmark averages. This fits the mechanism goal, and we provide curation logs, bootstrap CIs, and paired permutation tests with multiplicity correction to limit overclaiming. Where accuracy gaps lie within the per-combination paired-test noise floor (Appendix~\ref{app:statistics}), we report mechanism contrasts, not accuracy wins.

The positive accuracy signal concentrates on a single (model, regime) combination (\texttt{llama-4-scout} social). The ``unreliable but recoverable'' region of the boundary is therefore exhibited by one model in one regime, and the readiness gap remains an explanatory hypothesis. The claims we rest on are the mechanistic ones, the recovery account with its predictive test (Sections~\ref{sec:recovery_breakage} and~\ref{sec:predictive_analysis}) and the arbitrary-control baseline (Table~\ref{tab:random_action_main}), which hold across the matrix and on hard math. What would generalize the boundary claim is prospective validation on settings selected without reference to controller outcomes, in particular a second low-capability model reproducing the unreliable-but-recoverable pattern.

\paragraph{Scope of generalization.} Our conclusions are restricted to the evaluated setting: three open-weight model families served through one backend, homogeneous 3-agent teams with profile-differentiated prompts over a shared input, short-horizon collective judgment with 3--4 round budgets, and the specific task regimes studied. The regimes and boundary probes were selected as \emph{diagnostic} instruments that span round-0 reliability and recoverability, not as a representative sample of multi-agent workloads. The paper therefore makes no claim about how frequently the readiness-gap condition arises in naturally occurring deployments; establishing that base rate would require a workload survey we did not conduct. Because the boundary depends directly on round-0 reliability and recoverability, both of which shift with capability, it may sit elsewhere for stronger, weaker, or heterogeneous teams (mixed model families, differentiated roles, or asymmetric information), and we do not extrapolate to them.

The main study fixes 3-agent teams and 3--4 round budgets. The ablations in Appendix~\ref{app:team_size} and Appendix~\ref{app:round_budget} indicate the signatures are not artifacts of those choices. At 5 agents the transactional lock-in signature is invariant and the \texttt{llama-4-scout} social advantage of the exploration-oriented controllers persists, and at a round budget of 6 the main conclusions are unchanged. Larger teams and longer horizons remain unexplored. Situational control relies on count- and rationale-divergence signals, including lexical conflict markers, rather than entropy- or self-consistency-based uncertainty estimation. Such keyword-based triggers are auditable but rigid and may not transfer to free-form coordination settings. Richer signal designs, long-term trust calibration, repeated episodes, and human evaluation of perceived leadership quality are left for future work.
\section{Conclusion}
The contribution is to treat process-level coordination control as a \emph{contingency} question answered by measurement: not whether leadership controllers win, but under what conditions they add value, and whether those conditions match what team science predicts. Behavioral signatures (lock-in, exploration, recovery) replace single-number accuracy as the primary scientific object, and per-action ablations expose which controller components carry the effect in which regime. The measured boundary lines up with contingency theory: leadership substitutes, path-goal redundancy, and the situational readiness gap (Section~\ref{sec:theory_mapping}). A largely null accuracy result is the predicted outcome, not a failure of the controllers. The methodology is theory-grounded but theory-independent: leadership controllers are a first substantive case study, and the same measurement vocabulary applies to other team-science concepts (transactive memory, conflict regulation, trust calibration) treated as future case studies.

On the leadership case study, the methodology reveals structure beneath the controller labels: against a shared round-0 vote, transactional control reduces to the vote on all 12 main-matrix combinations, and the only significant accuracy gains fall on the one combination where an unreliable round-0 is also recoverable and not already repaired by plain interaction. That combination is where the readiness-gap hypothesis says a controller's interaction should earn its keep, and the regime our measurement vocabulary is built to detect; whether the hypothesis holds beyond the evaluated setting awaits prospective validation (Section~\ref{sec:limitations}).

\section*{Ethics Statement}
This paper studies coordination policies for multi-agent LLM teams on reasoning and social-judgment tasks. The main ethical risk is overinterpreting social or organizational metaphors as if they guaranteed desirable behavior. Our use of leadership terminology is operational rather than normative: we model leadership as explicit coordination control so that its behavioral consequences can be inspected, measured, and compared. A related risk is anthropomorphism: this language can encourage readers to over-attribute human-like intentions, social understanding, or organizational competence to the agents studied here. We do not claim that human leadership theories transfer wholesale to agent systems or that any single controller is universally beneficial.

The evaluation also includes social-norm and moral-judgment style data. Such datasets can reflect annotation artifacts, culturally narrow assumptions, or unstable judgments about acceptable behavior. To reduce that risk, we separate broad layers from narrower curated subsets, document fixed inclusion and exclusion rules, and report when effects weaken on broad layers.

The systems studied here are not intended for autonomous deployment in high-stakes settings. They are lightweight research controllers evaluated in short-horizon simulated team interactions. Any use of similar methods in consequential domains would require domain-specific validation, stronger uncertainty estimation, and substantially more careful human oversight than provided in the present study.
\section*{Reproducibility Statement}
We designed the paper to be reproducible at the level of data construction, execution protocol, and logged team behavior. All reported conditions share the same 3-agent execution loop, base prompt structure, stopping logic, and logging schema, with policy differences isolated to control decisions. We report regime-separated evaluations, broad layers, focused curated subsets, and a mixed benchmark so that readers can distinguish mechanism visibility from broader robustness. The same evaluation matrix is run across three open-weight model families (\texttt{gpt-oss-120b}, \texttt{llama-4-scout}, \texttt{gemma-4-31B-it}) through a single inference backend, allowing readers to separate controller-specific interaction signatures from model-family-specific ranking shifts. Because round~0 is generated once per (task, seed) and reused across conditions, the round-0-vote baseline is reproducible and identical across controllers.

The project includes scripts for dataset normalization, split construction, experiment execution, and behavioral-metric analysis. The paper also reports source composition, seed usage, and the key evaluation scales used in the final experiments. For curated subsets, we retain explicit construction rules and exclusion notes so that regime design remains auditable rather than implicit. \artifactstatement

Two practical limits remain. First, all three model families are open-weight models served through one gateway; coverage of proprietary and larger models is left to future work. Second, some controllers, especially the adaptive and dissent-preserving variants, remain lightweight policy implementations, not optimized systems. We therefore present this paper as a reproducible mechanism study rather than a maximally tuned benchmark submission.
\section*{LLM Usage Disclosure}
LLMs were used during the preparation of this work for limited writing and coding support, including editing prose, restructuring text, and helping implement experiment scripts. The key research ideas and scientific framing originated with the author. LLMs were not used to generate final scientific claims, to replace author judgment in data curation, or to serve as evaluation judges for the reported results. All reported analyses, interpretations, and final manuscript content were verified by the author.
\ifdefined\showacknowledgments
\section*{Acknowledgments}
The author acknowledges the Indiana University Pervasive Technology Institute \citep{stewart2017pti} for providing large-language-model inference resources, through the Research and Academic LLM Services (REALLMS) API, that have contributed to the research results reported within this paper. The author thanks the action editor and the three anonymous reviewers for their careful reading and constructive feedback.
\fi

\bibliographystyle{tmlr}
\bibliography{references}

\appendix
\section{Curation Protocol and Reporting Layers}
\label{app:curation_protocol}
This appendix documents the evaluation construction logic used throughout the paper. Our goal is not to optimize subsets for any single policy, but to isolate coordination regimes that are otherwise diluted by trivial items, annotation-sensitive items, or task formulations that do not induce the intended interaction pressure. For each regime, we therefore distinguish between a broad layer and a narrower curated layer used for more mechanism-focused analysis.

\begin{table*}[t]
\centering
\small
\resizebox{\textwidth}{!}{
\begin{tabular}{p{0.17\textwidth}p{0.21\textwidth}p{0.24\textwidth}p{0.15\textwidth}p{0.15\textwidth}}
\toprule
\textbf{Regime} & \textbf{Broad Layer} & \textbf{Construction Rule} & \textbf{Main Broad Scale} & \textbf{Curated Logic} \\
\midrule
Closed-ended shared-information QA & CommonsenseQA + StrategyQA reference pools & Preserve defended anchor items, extend with stable reference-pool items, balance sources 50/50, remove a small fixed list of benchmark-fragile or low-diagnostic items & 100 items & Smaller curated subsets retain only items with meaningful disagreement pressure and defensible gold labels \\
Abductive ambiguity & AlphaNLI stable reference pool & Start from the stable 100-item reference pool and apply no additional item-level filtering at 100-scale & 100 items & Curated subsets retain items where both hypotheses are plausible and causal ambiguity is preserved \\
Social-norm ambiguity & Social Chemistry broad \texttt{B/C} pool & Keep only \texttt{B/C} ternary labels, dev split, action-agree in [2.0, 3.0], and areas \texttt{confessions}, \texttt{amitheasshole}, \texttt{dearabby}; prioritize lower-agreement items and preserve area diversity & 100 items & Focused social cores retain items where context-dependent or unacceptable judgments remain interpretable without annotation artifacts \\
\bottomrule
\end{tabular}
}
\caption{Summary of the regime-construction protocol. Each broad layer is paired with a smaller curated set used to test whether the same qualitative policy pattern survives under tighter regime control.}
\label{tab:appendix_curation_summary}
\end{table*}

\subsection{Closed-Ended Shared-Information QA}
\label{app:closed_ended_regime}
The broad closed-ended regime is constructed from balanced CommonsenseQA and StrategyQA reference pools. The 100-item set preserves six defended anchor items from early pilot runs and then extends them with stable reference-pool items while maintaining a 50/50 source balance. The only exclusions are a small fixed list of benchmark-fragile or low-diagnostic items identified before the final conference-scale runs. This broad set is intended to test whether the early challenge-set pattern survives once the regime is widened toward a more verification-heavy shared-information workload.

\subsection{Abductive Ambiguity}
\label{app:alphanli_regime}
The AlphaNLI broad layer is intentionally simpler. At 100-scale, we use the stable reference pool directly and do not apply additional item-level exclusions. This makes AlphaNLI the cleanest anti-cherrypicking layer in the paper. The smaller curated AlphaNLI cores remain useful for qualitative and mechanism-focused analysis because they preserve especially strong explanatory ambiguity, but the 100-item broad set is already broad enough to support the paper's main quantitative conclusions.

\subsection{Social-Norm Ambiguity}
\label{app:social_regime}
The Social Chemistry broad layer uses a scalable ternary mapping and keeps only \texttt{B/C} judgments, where \texttt{B} denotes context-dependent social evaluation and \texttt{C} denotes unacceptable behavior. We exclude \texttt{A} items because they too often collapse into easy positive cases that do not require dissent handling. To reduce source-specific bias, the broad 100-item social set is balanced across three conflict-rich areas: \texttt{confessions}, \texttt{amitheasshole}, and \texttt{dearabby}. We further restrict to the dev split and to intermediate agreement levels (\texttt{action-agree} in [2.0, 3.0]) so that the social benchmark reflects genuine norm ambiguity rather than either trivial consensus or pure annotation noise.

\subsection{Why We Report Both Broad and Curated Layers}
The paper's main claim is about coordination mechanisms, not leaderboard optimization. That goal requires regimes in which the relevant coordination pressure is actually present. At the same time, curated regimes create a legitimate concern about post hoc selection. Reporting both layers addresses that concern directly. The broad layer answers the anti-cherrypicking question: does the qualitative pattern survive when we expand the regime? The curated layer answers the mechanism question: what happens when the regime is made maximally legible? We therefore treat broad and curated results as complementary rather than competing forms of evidence.
\section{Additional Robustness Results}

\subsection{Mixed-Benchmark Source Breakdown}
The mixed benchmark combines 25 CommonsenseQA items, 25 StrategyQA items, 50 AlphaNLI items, and 50 Social Chemistry items without exposing source identity to the controller. Table~\ref{tab:appendix_mixed_source_breakdown} reports the source-wise accuracy for all three models.

\begin{table*}[t]
\centering
\small
\resizebox{\textwidth}{!}{
\begin{tabular}{llccccc}
\toprule
\textbf{Model} & \textbf{Condition} & \textbf{AlphaNLI} & \textbf{CSQA} & \textbf{SocialChem} & \textbf{StratQA} & \textbf{Worst} \\
\midrule
\texttt{gpt-oss-120b} & Flat & 0.940 & 0.880 & 0.413 & 0.893 & 0.413 \\
 & Theory-free control & 0.913 & 0.840 & 0.360 & 0.867 & 0.360 \\
 & Transactional leadership & 0.940 & 0.880 & 0.393 & 0.840 & 0.393 \\
 & Transformational leadership & 0.913 & 0.853 & 0.373 & 0.813 & 0.373 \\
 & Situational leadership & 0.927 & 0.867 & 0.387 & 0.813 & 0.387 \\
\midrule
\texttt{llama-4-scout} & Flat & 0.873 & 0.773 & 0.420 & 0.827 & 0.420 \\
 & Theory-free control & 0.853 & 0.787 & 0.433 & 0.827 & 0.433 \\
 & Transactional leadership & 0.860 & 0.760 & 0.427 & 0.813 & 0.427 \\
 & Transformational leadership & 0.840 & 0.747 & 0.440 & 0.760 & 0.440 \\
 & Situational leadership & 0.847 & 0.773 & 0.513 & 0.733 & 0.513 \\
\midrule
\texttt{gemma-4-31B-it} & Flat & 0.913 & 0.787 & 0.353 & 0.787 & 0.353 \\
 & Theory-free control & 0.880 & 0.853 & 0.393 & 0.800 & 0.393 \\
 & Transactional leadership & 0.880 & 0.813 & 0.367 & 0.747 & 0.367 \\
 & Transformational leadership & 0.893 & 0.867 & 0.400 & 0.773 & 0.400 \\
 & Situational leadership & 0.887 & 0.840 & 0.413 & 0.747 & 0.413 \\
\bottomrule
\end{tabular}
}
\caption{Source-wise accuracy on the mixed 150-task benchmark (3 seeds) for all three models. The controllers differ mainly on the social slice, the hardest source; situational improves worst-source accuracy most on the weakest model (\texttt{llama-4-scout}, $0.513$ vs.\ transactional's $0.427$), the same regime where it gains over the round-0 vote.}
\label{tab:appendix_mixed_source_breakdown}
\end{table*}

The source-wise view explains the small aggregate gaps in Table~\ref{tab:main_results}: the controllers differ mainly on the social slice, the hardest source, where situational improves worst-source accuracy on \texttt{llama-4-scout} ($0.513$ vs.\ transactional's $0.427$). On the stronger models the slices compress and the controllers track each other. Behaviorally, transactional control keeps the highest mixed-workload lock-in and the fewest rounds on every model (Table~\ref{tab:appendix_main_stop_rounds}), while situational reopens more often, the same lock-in-vs-reopening tradeoff as on the separated regimes.

\subsection{Stopping-Round Comparison for the Main Regime Runs}
\label{app:main_stop_rounds}
\begin{table*}[t]
\centering
\small
\resizebox{\textwidth}{!}{
\begin{tabular}{llcccc}
\toprule
\textbf{Model} & \textbf{Condition} & Closed & AlphaNLI & Social & Mixed \\
\midrule
\texttt{gpt-oss-120b} & Flat & 1.40 & 1.16 & 1.57 & 1.26 \\
 & Theory-free control & 1.43 & 1.17 & 1.50 & 1.26 \\
 & Transactional leadership & 1.03 & 1.01 & 1.03 & 1.03 \\
 & Transformational leadership & 1.26 & 1.10 & 1.34 & 1.17 \\
 & Situational leadership & 1.21 & 1.19 & 1.39 & 1.21 \\
\midrule
\texttt{llama-4-scout} & Flat & 1.25 & 1.24 & 1.40 & 1.33 \\
 & Theory-free control & 1.29 & 1.23 & 1.48 & 1.38 \\
 & Transactional leadership & 1.00 & 1.00 & 1.03 & 1.02 \\
 & Transformational leadership & 1.18 & 1.15 & 1.29 & 1.24 \\
 & Situational leadership & 1.30 & 1.34 & 1.45 & 1.37 \\
\midrule
\texttt{gemma-4-31B-it} & Flat & 1.25 & 1.11 & 1.39 & 1.22 \\
 & Theory-free control & 1.30 & 1.10 & 1.36 & 1.20 \\
 & Transactional leadership & 1.00 & 1.00 & 1.00 & 1.00 \\
 & Transformational leadership & 1.19 & 1.06 & 1.25 & 1.14 \\
 & Situational leadership & 1.21 & 1.27 & 1.28 & 1.24 \\
\bottomrule
\end{tabular}
}
\caption{Mean stopping round across the four regimes (3 seeds). Transactional control stops at round~0 in nearly every episode (Stop@R $\approx 1.0$), while situational and transformational spend extra rounds reopening deliberation.}
\label{tab:appendix_main_stop_rounds}
\end{table*}

\subsection{Token Cost}
\label{app:tokens}
Table~\ref{tab:appendix_tokens} reports mean prompt-plus-completion tokens per run, pooled over the four regimes. Transactional control and its accept-only ablation are cheapest because they stop at round~0; flat and situational spend the most, mirroring the stopping-round ordering (Table~\ref{tab:appendix_main_stop_rounds}).

\begin{table}[h]
\centering
\small
\begin{tabular}{lccc}
\toprule
\textbf{Condition} & \texttt{gpt-oss-120b} & \texttt{llama-4-scout} & \texttt{gemma-4-31B-it} \\
\midrule
Flat & 1713 & 1082 & 839 \\
Transactional & 1197 & 658 & 583 \\
\quad\emph{accept-only} & 1153 & 649 & 582 \\
Transformational & 1438 & 816 & 683 \\
Situational & 1507 & 923 & 749 \\
\bottomrule
\end{tabular}
\caption{Mean tokens per run (prompt + completion), pooled over the four main regimes. Transactional and its accept-only ablation are cheapest (round-0 stop); flat and situational are the most expensive.}
\label{tab:appendix_tokens}
\end{table}

\subsection{Inclusive Leadership on the Social Regime}
\label{app:focused_social_core}
We include inclusive leadership (a dissent-preservation specialization of the control vocabulary that explicitly requests and records minority objections before synthesis) as a probe on the broad social regime (3 seeds; an independent run with its own shared round~0, whose main-condition accuracies differ from Table~\ref{tab:main_results} by at most $\sim$2pp). Accuracy is $0.457$ / $0.467$ / $0.403$ for inclusive on \texttt{gpt-oss-120b} / \texttt{llama-4-scout} / \texttt{gemma-4-31B-it}, against situational's $0.437$ / $0.517$ / $0.397$ and flat's $0.413$ / $0.420$ / $0.383$. Inclusive is competitive with situational on the stronger models, but on \texttt{llama-4-scout}, where the round-0 majority is least reliable, situational's active reopening ($0.517$) outperforms inclusive's dissent-preservation ($0.467$). Dissent preservation is a real mechanism, but realizing an accuracy gain still requires reopening the unreliable majority, not merely recording the minority view.

\subsection{Execution Settings}
\label{app:execution_settings}
The released configs make the execution settings explicit. All reported runs use 3-agent teams, the shared episode/round execution loop with a generated-once shared round~0 (Section~\ref{sec:experimental_setup}), and task-type round budgets of 3 rounds for closed-ended tasks and 4 for integration-style tasks. All three models (\texttt{gpt-oss-120b}, \texttt{llama-4-scout}, \texttt{gemma-4-31B-it}) are served through the same self-hosted backend with \texttt{temperature}=0.35. Generation length is \texttt{max\_tokens}=512 on the main matrix; the MATH-500 Level~5 evaluations use a 3-round budget for open-ended numeric tasks and \texttt{max\_tokens}=1024, raised to 3072 for \texttt{gpt-oss-120b} (a reasoning model) so that its round-0 reasoning is not truncated before it commits a final answer. The experiment code sets no explicit top-\emph{p} override or context-window cap, leaving provider defaults in force on those dimensions. The canonical configs for every reported run are the shared-round-0 family (\texttt{sr0\_*.json} in the released artifact), whose values match the settings stated here. Token-cost figures are computed from the per-call token counts logged in the run traces, not from the generation cap.

\subsection{Statistical Tests Across All Combinations}
\label{app:statistics}
For every (model, regime, condition) combination we report bootstrap 95\% CIs over item-level scores ($10^4$ iters) and paired permutation tests over per-task scores ($10^4$ iters, pairing on common (task, seed)) for every pairwise condition comparison; the complete table is released as a supplementary file. Table~\ref{tab:appendix_pairwise} summarizes representative highlights, recomputed on the shared-round-0 runs.

\begin{table}[h]
\centering
\small
\begin{tabular}{lllcc}
\toprule
\textbf{Model / Regime} & \textbf{Pair (A vs.\ B)} & \textbf{Metric} & \textbf{Diff} & \textbf{$p$} \\
\midrule
\texttt{llama-4-scout} / social & situational vs.\ transactional & acc & $+0.083$ & $0.004$ \\
\texttt{llama-4-scout} / social & transformational vs.\ transactional & acc & $+0.053$ & $0.007$ \\
\texttt{llama-4-scout} / social & situational vs.\ flat & acc & $+0.080$ & $0.007$ \\
\texttt{gpt-oss-120b} / social & situational vs.\ transactional & acc & $+0.027$ & $0.113$ \\
\texttt{llama-4-scout} / mixed & transactional vs.\ situational & lock-in & $+0.124$ & $<0.0001$ \\
\texttt{gpt-oss-120b} / mixed & transactional vs.\ situational & lock-in & $+0.058$ & $<0.0001$ \\
\texttt{gemma-4-31B-it} / mixed & transactional vs.\ situational & lock-in & $+0.064$ & $<0.0001$ \\
\bottomrule
\end{tabular}
\caption{Representative paired-permutation contrasts on the shared-round-0 runs ($10^4$ iters). Accuracy gains concentrate on \texttt{llama-4-scout} social (the unreliable-round-0 combination); lock-in differences between transactional and the other controllers are significant in every (model, regime) combination, confirming the behavioral signature is more stable than the accuracy ordering.}
\label{tab:appendix_pairwise}
\end{table}

\subsection{Reporting Philosophy}
We include these appendix tables because the paper's main risk is not underpowered experimentation; it is overinterpretation of carefully designed regimes. Our solution is to make the regime design itself auditable. Broad layers, source-wise mixed breakdowns, process metrics, and paired statistical tests together provide a transparent trail from raw workload construction to the mechanism claims made in the main text.

\subsection{MAD Baseline and Situational Random-Switch Ablation}
\label{app:mad_siturand}
We benchmark the leadership controllers against a stronger peer-to-peer baseline on the \emph{full} main matrix, and separately isolate the signal-driven component of situational control. \emph{MAD} is a peer-to-peer multi-agent debate baseline following \citet{du2024improving}: each round agents read peer answers and may revise, with no leader and no acceptance criterion; the final answer is a majority vote. We run MAD against \texttt{flat}, \texttt{situational}, and \texttt{transformational} on all 12 (model, regime) combinations on the same generated-once shared round~0 (3 seeds, Table~\ref{tab:appendix_mad_full}); these are independent runs from the main matrix whose accuracies differ by at most $\sim$2pp.

\begin{table}[h]
\centering
\small
\begin{tabular}{llccccc}
\toprule
\textbf{Model} & \textbf{Regime} & R0-vote & Flat & MAD & Situ. & Transf. \\
\midrule
\texttt{gpt-oss-120b} & AlphaNLI & 0.937 & 0.930 & 0.923 & 0.927 & 0.907 \\
 & Closed & 0.863 & 0.870 & 0.877 & 0.857 & 0.857 \\
 & Social & 0.413 & 0.410 & 0.410 & 0.427 & 0.430 \\
 & Mixed & 0.713 & 0.724 & 0.722 & 0.716 & 0.718 \\
\midrule
\texttt{llama-4-scout} & AlphaNLI & 0.870 & 0.870 & 0.873 & 0.897 & 0.883 \\
 & Closed & 0.790 & 0.797 & 0.783 & 0.790 & 0.803 \\
 & Social & 0.437 & 0.453 & 0.450 & \textbf{0.513} & 0.467 \\
 & Mixed & 0.693 & 0.711 & 0.713 & 0.707 & 0.696 \\
\midrule
\texttt{gemma-4-31B-it} & AlphaNLI & 0.927 & 0.937 & 0.930 & 0.920 & 0.917 \\
 & Closed & 0.777 & 0.797 & 0.807 & 0.777 & 0.777 \\
 & Social & 0.390 & 0.390 & 0.377 & 0.400 & 0.387 \\
 & Mixed & 0.678 & 0.671 & 0.680 & 0.696 & 0.691 \\
\bottomrule
\end{tabular}
\caption{Full-matrix MAD baseline (3 seeds, shared round~0). MAD never \emph{significantly} beats a leadership controller; where it nominally edges them the margin is $\leq 3$pp on reliable-round-0 regimes where no condition separates from the vote. On the one combination with a real gain, \texttt{llama-4-scout} social, MAD ($0.450$) tracks flat ($0.453$) and the vote ($0.437$) and loses to situational ($0.513$) by $6.3$pp ($p=0.01$).}
\label{tab:appendix_mad_full}
\end{table}

The full-matrix comparison sharpens the contingency picture: MAD, the recognized peer-debate baseline, behaves like undirected \texttt{flat} interaction everywhere. It clusters at the round-0 ceiling on the reliable regimes (its $\leq 3$pp nominal edges over a controller are never significant and fall where nothing separates), and it \emph{fails to capture the readiness-gap recovery} on \texttt{llama-4-scout} social, where the signal-driven situational controller beats it by $6.3$pp. Generic peer debate is not a substitute for the signal-driven trigger.

\paragraph{Situational random-switch.} To isolate that trigger, we replace situational's team-state signals with a fixed exploration probability $p=0.45$ (matched to the empirical exploration rate) while keeping its control space, on AlphaNLI / social / mixed (3 seeds; an independent run, $\sim$2pp drift). On \texttt{llama-4-scout} social (the one unreliable-round-0 combination), signal-driven situational ($0.517$) beats random-switch ($0.477$) and flat ($0.430$), whereas on the reliable regimes the variants compress (\texttt{gpt-oss-120b}: $0.430$ vs $0.417$; \texttt{gemma-4-31B-it}: $0.393$ vs $0.390$). The advantage comes from \emph{when} the controller reopens exploration, not from the larger action space, the same round-0-reliability pattern as the main matrix.

\subsection{Arbitrary-Control Baseline}
\label{app:random_action}
To test whether the recovery the leadership controllers produce comes from the theory-derived decision rule or merely from access to the control vocabulary, we add an \emph{arbitrary-control} baseline: each round it assigns every agent a uniformly random action from \{\texttt{accept}, \texttt{revise}, \texttt{explore}\}, with no leadership-derived structure. It shares the action set of the leadership controllers but its decision rule is chance. We run it alongside flat, transactional, transformational, and situational control on the AlphaNLI, social, and mixed regimes across the three models (3 seeds, shared round~0). The pooled recovery and breakage rates over these nine combinations are reported in the main text (Table~\ref{tab:random_action_main}, Section~\ref{sec:recovery_breakage}): the arbitrary controller's recovery ($0.009$) is indistinguishable from transactional control ($0.010$), which is itself the round-0 vote, and below flat ($0.061$). Random choices are keyed deterministically by (run, round, agent), making the baseline reproducible; \texttt{accept} keeps the agent's current answer without regeneration, while \texttt{revise} and \texttt{explore} re-prompt the agent as in the leadership controllers. The theory-derived controllers' recovery (situational $0.133$, transformational $0.089$) therefore reflects what the theory contributes, a decision rule for composing the action set, not the action set alone.

\subsection{Case Studies: Recovery, Failed Recovery, and Breakage}
\label{app:case_studies}
The recovery and breakage rates of Section~\ref{sec:recovery_breakage} are pooled statistics; this appendix shows what the underlying trajectories look like. All three cases come from the same controller and combination (situational on \texttt{llama-4-scout} social, seed 0) and thus form one controlled contrast. They are drawn from pools of 49 recovery, 123 failed-recovery, and 23 breakage episodes for this combination; the pool ratios reproduce the reported rates (recovery $49/172=0.285$, breakage $0.18$), and the examples are therefore representative of the mechanism, not curated exceptions.

\paragraph{Case 1: successful recovery (correct minority survives reopening).}
Item: judging ``making people do things they don't care to do'' when a partner drags the author to a movie (gold: unacceptable). Round-0 independent answers are \{C, A, A\}: the round-0 majority is A (acceptable) and wrong, while the gold answer C is a one-agent minority. The situational trigger reads the 2--1 split and issues \texttt{explore} to all three agents; in round 1 the answers move to \{C, B, C\} and the final answer is C, correct. The correct round-0 minority survived reopening and spread, which is the dissent-preservation route by which recovery happens.

\paragraph{Case 2: failed recovery (correct answer present but lost).}
Item: judging ``talking back to your parents'' when standing up to one's father (gold: unacceptable). Round-0 answers are \{A, B, C\}, one of each: the majority-by-tiebreak is wrong, and the gold answer C is again in the room. Reopening moves round 1 to \{B, B, B\} and the final answer is B, wrong. This is the mirror image of Case 1: reopening converged on the middling compromise, not the correct minority. Recovery is not guaranteed even when the correct answer exists at round 0, which is why the recovery rate is only $0.285$.

\paragraph{Case 3: breakage (correct consensus reopened to error).}
Item: judging ``feeding your pets leftovers'' against a vet's advice (gold: unacceptable). Round-0 answers are unanimous \{C, C, C\} and correct. The controller nonetheless issues \texttt{explore}, and round 1 moves to \{B, B, B\}: the final answer is B, wrong. The reopening on a nominally unanimous round 0 is the documented trigger, not an artifact: situational reopens when the rationales diverge (three distinct rationale openings plus hedged conflict markers from two or more agents, which these rationales contain) even if the answer labels agree. This case makes the breakage cost concrete. The same trigger that repairs wrong majorities occasionally drags a correct consensus to error, which is exactly the brittleness the recovery/breakage decomposition quantifies.

\subsection{Unconstrained-Coordinator Baseline}
\label{app:llm_coordinator}
The arbitrary-control baseline (Appendix~\ref{app:random_action}) tests the action vocabulary without a principled rule. A complementary question is whether any \emph{explicit structure} is needed at all: would a plain LLM, acting as the team's coordinator with no leadership principles embedded, reproduce what the theory-derived controllers do? We add an \emph{unconstrained-coordinator} baseline to answer it. Each round, a coordinator instance of the same model reads the question and the team's current answers and writes free-form guidance to each agent in its own words, plus a continue/stop decision (the minimal protocol needed to terminate an episode). It has no action vocabulary, no lexical or threshold decision rule, and no style-derived structure. To keep the comparison confound-free, every episode starts from the identical logged shared round~0 of the original matrix (injected verbatim), and the final aggregation is the same majority rule as every other condition, so that any difference is attributable to the coordinator's rounds-1+ interaction. The coordinator's own generation calls are logged with token counts and counted in cost.

We note that \texttt{llama-4-scout} was removed from our self-hosted backend after the original runs, and that its coordinator-baseline rounds 1+ are therefore served through a single pinned external provider at the same temperature and generation budget, with round~0 still injected from the original self-hosted logs. The other two models run entirely on the original backend. We flag the affected rows in Table~\ref{tab:appendix_coordinator}.

\begin{table}[h]
\centering
\small
\setlength{\tabcolsep}{4.5pt}
\begin{tabular}{llccccccc}
\toprule
\textbf{Model} & \textbf{Regime} & \textbf{R0-vote} & \textbf{Coord.} & $\Delta$vote & $\Delta$flat & $\Delta$situ.\ & Recov.\ & Break.\ \\
\midrule
\texttt{gpt-oss-120b} & AlphaNLI & 0.923 & 0.920 & $-0.003$ & $+0.000$ & $-0.003$ & 0.130 & 0.014 \\
\texttt{gpt-oss-120b} & Closed   & 0.865 & 0.859 & $-0.007$ & $-0.007$ & $-0.007$ & 0.175 & 0.035 \\
\texttt{gpt-oss-120b} & Social   & 0.423 & 0.427 & $+0.003$ & $+0.010$ & $-0.023$ & 0.064 & 0.079 \\
\texttt{gpt-oss-120b} & Mixed    & 0.736 & 0.729 & $-0.007$ & $-0.018$ & $+0.011$ & 0.050 & 0.027 \\
\texttt{llama-4-scout}\textsuperscript{\dag} & AlphaNLI & 0.867 & 0.897 & $+0.030$ & $+0.023$ & $+0.003$ & 0.300 & 0.012 \\
\texttt{llama-4-scout}\textsuperscript{\dag} & Closed   & 0.780 & 0.770 & $-0.010$ & $-0.017$ & $-0.013$ & 0.106 & 0.043 \\
\texttt{llama-4-scout}\textsuperscript{\dag} & Social   & 0.427 & 0.453 & $+0.027$ & $+0.020$ & $\mathbf{-0.060}$\sigstar & 0.140 & 0.125 \\
\texttt{llama-4-scout}\textsuperscript{\dag} & Mixed    & 0.693 & 0.724 & $+0.031$ & $+0.027$ & $+0.020$ & 0.246 & 0.064 \\
\texttt{gemma-4-31B-it} & AlphaNLI & 0.930 & 0.873 & $\mathbf{-0.057}$\sigstar & $\mathbf{-0.077}$\sigstar & $\mathbf{-0.043}$\sigstar & 0.000 & 0.061 \\
\texttt{gemma-4-31B-it} & Closed   & 0.760 & 0.693 & $\mathbf{-0.067}$\sigstar & $\mathbf{-0.100}$\sigstar & $\mathbf{-0.090}$\sigstar & 0.125 & 0.127 \\
\texttt{gemma-4-31B-it} & Social   & 0.387 & 0.303 & $\mathbf{-0.083}$\sigstar & $\mathbf{-0.080}$\sigstar & $\mathbf{-0.077}$\sigstar & 0.016 & 0.241 \\
\texttt{gemma-4-31B-it} & Mixed    & 0.676 & 0.609 & $\mathbf{-0.067}$\sigstar & $\mathbf{-0.076}$\sigstar & $\mathbf{-0.089}$\sigstar & 0.021 & 0.109 \\
\bottomrule
\end{tabular}
\caption{Unconstrained-coordinator baseline on the 12 main-matrix combinations (3 seeds; identical injected shared round~0). $\Delta$vote, $\Delta$flat, and $\Delta$situ.\ are the coordinator's accuracy minus the round-0 vote, flat, and situational control respectively (paired permutation tests); \sigstar\ marks differences significant after Benjamini--Hochberg correction within each comparison family. Recov.\ and Break.\ are the coordinator's recovery and breakage. \textsuperscript{\dag}Rounds 1+ served through a pinned external provider (see text).}
\label{tab:appendix_coordinator}
\end{table}

Three patterns emerge. First, where the round-0 majority is reliable (\texttt{gpt-oss-120b}), the coordinator reduces to the round-0 vote (every $|\Delta\text{vote}| \le 0.7$pp, none significant). Reading agreeing answers, it stops almost immediately (mean stopping round $1.1$--$1.4$), reproducing the leadership-substitutes region without any explicit rule. Second, on \texttt{gemma-4-31B-it}, the model that cannot repair its own errors, the coordinator is significantly \emph{harmful} in all four regimes ($-5.7$ to $-8.3$pp against the vote and $-7.6$ to $-10.0$pp against flat, all $q<0.01$). Its interventions produce breakage (up to $0.241$ on the social regime) with near-zero recovery ($0.016$), which is the same brittleness of unprincipled reopening that the arbitrary-control baseline shows in miniature. Third, on \texttt{llama-4-scout}, the recoverable low-capability model, the coordinator gains a little but not significantly (at most $+3.1$pp, all $q\ge 0.08$), and on the readiness-gap combination (social) it stays far below situational control: $-6.0$pp against situational on the same items (paired $p=0.016$), against situational's $+8.7$pp over the shared vote. An unconstrained coordinator therefore does not reproduce the theory-derived controllers: where coordination is unnecessary it is merely inert, where the team cannot recover it is actively destructive, and in the one region where directed reopening pays it captures little of the available gain. Access to a coordinator, even one free to say anything, is not a substitute for a principled decision rule about when to reopen and what to direct.

\subsection{Team-Size Ablation}
\label{app:team_size}
We extend the AlphaNLI and social regimes to $n=5$ agents on all three models (3 seeds, $n=300$ per condition) to test whether the controller signatures persist beyond the 3-agent main study (Table~\ref{tab:appendix_team5}).

\begin{table}[h]
\centering
\small
\begin{tabular}{llcccccc}
\toprule
 & & \multicolumn{3}{c}{\textbf{AlphaNLI}} & \multicolumn{3}{c}{\textbf{Social}} \\
\cmidrule(lr){3-5} \cmidrule(lr){6-8}
\textbf{Model} & \textbf{Condition} & Acc & Lock-In & Explore & Acc & Lock-In & Explore \\
\midrule
\texttt{gpt-oss-120b} & Flat & 0.900 & 0.920 & 0.000 & 0.423 & 0.913 & 0.000 \\
 & Transactional & 0.933 & \textbf{1.000} & 0.000 & 0.447 & \textbf{1.000} & 0.000 \\
 & Transformational & 0.913 & 0.960 & 0.743 & 0.437 & 0.897 & 0.857 \\
 & Situational & 0.933 & 1.000 & 0.090 & 0.420 & 0.960 & 0.170 \\
\midrule
\texttt{llama-4-scout} & Flat & 0.883 & 0.947 & 0.000 & 0.440 & 0.883 & 0.000 \\
 & Transactional & 0.863 & \textbf{1.000} & 0.000 & 0.433 & \textbf{1.000} & 0.000 \\
 & Transformational & 0.893 & 0.943 & 0.400 & \textbf{0.503} & 0.817 & 0.513 \\
 & Situational & 0.860 & 0.997 & 0.300 & 0.493 & 0.873 & 0.217 \\
\midrule
\texttt{gemma-4-31B-it} & Flat & 0.953 & 0.940 & 0.000 & 0.360 & 0.923 & 0.000 \\
 & Transactional & 0.920 & \textbf{1.000} & 0.000 & 0.393 & \textbf{1.000} & 0.000 \\
 & Transformational & 0.930 & 0.990 & 0.840 & 0.360 & 0.887 & 0.720 \\
 & Situational & 0.913 & 0.993 & 0.090 & 0.390 & 0.960 & 0.120 \\
\bottomrule
\end{tabular}
\caption{Team-size $=5$ ablation (3 seeds, $n=300$ per condition). The transactional lock-in signature stays invariant ($1.000$ on every model) with zero exploration, while transformational and situational explore and lower lock-in. On \texttt{llama-4-scout} social (the unreliable-round-0 combination), the transformational/situational accuracy advantage over flat and transactional persists at $n=5$, confirming the round-0-reliability finding is not an artifact of the 3-agent team size.}
\label{tab:appendix_team5}
\end{table}

\subsection{Round-Budget Ablation}
\label{app:round_budget}
We re-evaluate the AlphaNLI and social regimes with \texttt{round\_budget}=6 (instead of the 3--4 used in the main runs) on all three models, to verify that the main results do not depend on the round-budget cap and that exploration-oriented controllers are not disadvantaged by it (Table~\ref{tab:appendix_round6}; 3 seeds, $n=300$ per condition).

\begin{table}[h]
\centering
\small
\begin{tabular}{llcc}
\toprule
\textbf{Model} & \textbf{Condition} & AlphaNLI & Social \\
\midrule
\texttt{gpt-oss-120b} & Flat & 0.913 & 0.427 \\
 & Transactional & 0.923 & 0.423 \\
 & Transformational & 0.903 & 0.440 \\
 & Situational & 0.900 & 0.437 \\
\midrule
\texttt{llama-4-scout} & Flat & 0.893 & 0.413 \\
 & Transactional & 0.867 & 0.440 \\
 & Transformational & 0.887 & 0.477 \\
 & Situational & 0.880 & \textbf{0.527} \\
\midrule
\texttt{gemma-4-31B-it} & Flat & 0.940 & 0.383 \\
 & Transactional & 0.927 & 0.393 \\
 & Transformational & 0.923 & 0.377 \\
 & Situational & 0.927 & 0.383 \\
\bottomrule
\end{tabular}
\caption{Round-budget $=6$ ablation (3 seeds, $n=300$ per condition), accuracy by regime. The cap is not binding, transactional still stops at round~0 (as in the main runs, Table~\ref{tab:appendix_main_stop_rounds}), and the accuracy ranking is preserved. On \texttt{llama-4-scout} social, the one unreliable-round-0 combination, situational \emph{gains} from the larger budget ($0.527$ vs.\ $0.513$ at budget~3), consistent with reopening helping more when more incorrect round-0 majorities remain to repair.}
\label{tab:appendix_round6}
\end{table}

\subsection{Boundary probes: full results}
\label{app:boundary_probes}
These are the four regimes used in Section~\ref{sec:recovery_breakage} to test where a controller can beat plain interaction, reported here in full. All use the same shared-round-0 protocol, 3-agent teams, five conditions, and three seeds as the main matrix. ANLI-R3 \citep{nie2020adversarial} is the most adversarial round of Adversarial NLI (balanced three-way entailment/neutral/contradiction, $n=100$; Table~\ref{tab:appendix_anli}); Winogrande \citep{sakaguchi2021winogrande} is the twin-sentence coreference benchmark ($n=100$, binary; Table~\ref{tab:appendix_winogrande}). Both have objective gold, and the 3-agent round-0 majority stays reliable (round-0 vote $0.70$--$0.90$) even though individual agents are often fooled. There is therefore little incorrect consensus to repair, and no controller separates from flat. The Scruples regime \citep{lourie2021scruples} is built from \texttt{r/AmITheAsshole} anecdotes: we keep only \emph{contested} posts (plurality of community votes below $0.7$ over at least five votes), balanced across a three-way verdict (author at fault / not at fault / mixed-or-unclear), $n=99$, median community agreement $0.56$ (Table~\ref{tab:appendix_scruples}). Here the round-0 majority is genuinely unreliable (vote $0.39$--$0.49$) but unrecoverable: predicting the crowd's plurality on a contested moral case is not derivable from the text by reasoning. Recovery is thus near zero for every controller (including flat), and again no controller beats flat. Only the social regime of the main matrix combines an unreliable round-0 with a recoverable signal that flat does not already exploit (Table~\ref{tab:boundary}).

\begin{table}[h]
\centering
\small
\begin{tabular}{lccc}
\toprule
\textbf{Condition} & \texttt{gpt-oss-120b} & \texttt{llama-4-scout} & \texttt{gemma-4-31B-it} \\
\midrule
Round-0 vote & 0.697 & 0.730 & 0.717 \\
\midrule
Flat & 0.660 & 0.767 & 0.723 \\
Theory-free control & 0.683 & 0.707 & 0.737 \\
Transactional & 0.700 & 0.733 & 0.717 \\
Transformational & 0.673 & 0.693 & 0.740 \\
Situational & 0.673 & 0.703 & 0.727 \\
\bottomrule
\end{tabular}
\caption{ANLI-R3 (adversarial NLI), accuracy, $n=100$, 3 seeds. Objective gold; the round-0 majority is reliable and no controller separates from flat.}
\label{tab:appendix_anli}
\end{table}

\begin{table}[h]
\centering
\small
\begin{tabular}{lccc}
\toprule
\textbf{Condition} & \texttt{gpt-oss-120b} & \texttt{llama-4-scout} & \texttt{gemma-4-31B-it} \\
\midrule
Round-0 vote & 0.843 & 0.783 & 0.900 \\
\midrule
Flat & 0.860 & 0.787 & 0.913 \\
Theory-free control & 0.853 & 0.810 & 0.910 \\
Transactional & 0.847 & 0.783 & 0.900 \\
Transformational & 0.857 & 0.780 & 0.897 \\
Situational & 0.843 & 0.780 & 0.893 \\
\bottomrule
\end{tabular}
\caption{Winogrande (twin-sentence coreference), accuracy, $n=100$, 3 seeds. Objective gold; the round-0 majority is reliable and no controller separates from flat.}
\label{tab:appendix_winogrande}
\end{table}

\begin{table}[h]
\centering
\small
\begin{tabular}{lccc}
\toprule
\textbf{Condition} & \texttt{gpt-oss-120b} & \texttt{llama-4-scout} & \texttt{gemma-4-31B-it} \\
\midrule
Round-0 vote & 0.451 & 0.391 & 0.485 \\
\midrule
Flat & 0.461 & 0.394 & 0.492 \\
Theory-free control & 0.451 & 0.394 & 0.488 \\
Transactional & 0.451 & 0.391 & 0.481 \\
Transformational & 0.448 & 0.394 & 0.495 \\
Situational & 0.461 & 0.384 & 0.481 \\
\bottomrule
\end{tabular}
\caption{Scruples contested moral judgments, accuracy, $n=99$, 3 seeds. Subjective/contested gold; the round-0 majority is unreliable (vote $0.39$--$0.49$) but unrecoverable, and every controller, including flat, stays at the vote.}
\label{tab:appendix_scruples}
\end{table}

\subsection{Situational trigger sensitivity}
\label{app:situational_sensitivity}
The situational controller's reopening rule uses fixed thresholds and a hand-built conflict-marker lexicon (Section~\ref{sec:method}). To test whether the social-regime gain depends on those choices, we re-run the social regime with three perturbed variants alongside the main controller, all sharing round-0 (an independent run with its own generated-once round~0; accuracies differ from the main matrix by at most $\sim$2pp): \emph{split-3-only} drops the 2--1-split trigger and reopens only on a full three-way disagreement; \emph{lenient} relaxes the unanimous-but-divergent branch (rationale openings $\geq 2$, conflict markers from $\geq 1$ agent); and \emph{alt-lexicon} swaps in a different but comparable conflict-marker word list. Table~\ref{tab:appendix_sensitivity} reports accuracy, gain over flat, and recovery.

\begin{table}[h]
\centering
\small
\begin{tabular}{llccc}
\toprule
\textbf{Model} & \textbf{Situational variant} & Acc. & $\Delta$ vs flat & Recov. \\
\midrule
\texttt{gpt-oss-120b} & (flat / round-0 vote 0.427) & 0.437 & n/a & 0.070 \\
 & main & 0.420 & $-0.017$ & 0.047 \\
 & split-3-only & 0.433 & $-0.003$ & 0.023 \\
 & lenient & 0.437 & $+0.000$ & 0.064 \\
 & alt-lexicon & 0.440 & $+0.003$ & 0.064 \\
\midrule
\texttt{llama-4-scout} & (flat / round-0 vote 0.433) & 0.450 & n/a & 0.065 \\
 & main & 0.527 & $+0.077$\sigstar & 0.282 \\
 & split-3-only & 0.457 & $+0.007$ & 0.071 \\
 & lenient & 0.540 & $+0.090$\sigstar & 0.329 \\
 & alt-lexicon & 0.497 & $+0.047$ & 0.218 \\
\midrule
\texttt{gemma-4-31B-it} & (flat / round-0 vote 0.380) & 0.383 & n/a & 0.027 \\
 & main & 0.400 & $+0.017$ & 0.059 \\
 & split-3-only & 0.390 & $+0.007$ & 0.016 \\
 & lenient & 0.400 & $+0.017$ & 0.065 \\
 & alt-lexicon & 0.410 & $+0.027$ & 0.075 \\
\bottomrule
\end{tabular}
\caption{Situational sensitivity on the social regime (3 seeds; independent shared-round-0 run in which both accuracy and recovery drift modestly from the main matrix, e.g.\ flat recovery $0.065$ here vs.\ $0.04$ there, and within-run contrasts are the comparison of interest). $\Delta$ vs flat is paired-permutation; \sigstar\ marks a significant gain $\geq 5$pp. On \texttt{llama-4-scout} the gain survives the lexicon swap ($+4.7$pp, recovery $0.22$) and \emph{grows} under threshold relaxation ($+9.0$pp), which rules out an artifact of the markers or cutoffs; it collapses only under \emph{split-3-only}, which removes the 2--1-split trigger and with it the recovery ($0.28\rightarrow0.07$). \texttt{gpt-oss-120b} and \texttt{gemma-4-31B-it} show no gain under any variant.}
\label{tab:appendix_sensitivity}
\end{table}

First, the gain is \emph{not} an artifact of the hand-built lexicon or the precise thresholds: it survives swapping the conflict-marker list and grows when the thresholds are relaxed. Second, it is mechanistically localized to one trigger: reopening 2--1 majority splits. Removing that trigger (\emph{split-3-only}) drops recovery from $0.28$ to $0.07$ and the gain to near zero, identifying reopening of thin, contested majorities (not the lexicon or the fine thresholds) as the active ingredient, which is precisely the readiness-gap intervention of Section~\ref{sec:theory_mapping}.

\section{Policy Pseudocode}
\label{app:policy_pseudocode}
This appendix makes the controller logic more explicit. The goal is not to claim that the current implementations are the only valid operationalizations of these leadership theories, but to show the concrete decision flow used in the experiments. The first three subsections cover the three main controllers; the fourth, inclusive leadership, is the dissent-preservation probe of Appendix~\ref{app:focused_social_core}, not an additional main controller.

\subsection{Transactional Leadership}
\begin{quote}\ttfamily
Input: task $x$, round-$t$ answers $Y_t$ \\
$m \leftarrow$ majority answer among non-empty $Y_t$ \\
for each agent response $y_{i,t}$: \\
\hspace*{1em}if $y_{i,t}$ empty or $y_{i,t} \neq m$: issue REVISE \\
\hspace*{1em}else ($y_{i,t} = m$): issue ACCEPT \\
if majority count $\geq 2$ (2-of-3): \\
\hspace*{1em}SYNTHESIZE final answer from accepted pool \\
else: \\
\hspace*{1em}continue to next round
\end{quote}

The acceptance criterion is concrete: an agent's answer is accepted if it matches the current-round majority answer and revised otherwise; the team closes the round on a 2-of-3 majority. The defining property is strong convergence pressure: disagreement is treated primarily as a signal for correction rather than for further exploration.

\subsection{Transformational Leadership}
\begin{quote}\ttfamily
Input: task $x$, round-$t$ answers $Y_t$, team goal $g_t$ \\
broadcast shared objective $g_t$ to the team \\
for each agent: \\
\hspace*{1em}assign differentiated exploration directive $d_{i,t}$ \\
\hspace*{1em}request alternative framing, evidence path, or hypothesis \\
if distinct plausible lines remain active: \\
\hspace*{1em}delay synthesis and preserve diversity one more round \\
else: \\
\hspace*{1em}SYNTHESIZE final answer
\end{quote}

The defining property is structured diversity around a common team objective rather than local correction toward immediate agreement.

\subsection{Situational Leadership}
\begin{quote}\ttfamily
Input: task $x$, round-$t$ answers $Y_t$ (3 agents) \\
$u \leftarrow$ \# distinct non-empty answers \\
$d \leftarrow$ \# distinct rationale openings (first-sentence prefixes) \\
$c \leftarrow$ \# agents whose text contains a conflict marker \\
if $u \geq 3$ (three-way split): open EXPLORE \\
elif $u = 2$ (2--1 split): open EXPLORE \\
elif $u = 1$ and $d \geq 3$ and $c \geq 2$: open EXPLORE \\
else: fall back to transactional-style convergence \\
\hspace*{1em}(REVISE / ACCEPT, then SYNTHESIZE on a 2-of-3 majority)
\end{quote}

The conflict markers are a fixed lexicon (``however,'' ``but,'' ``although,'' ``contradict,'' ``unlikely,'' ``fails,'' and similar). The defining property is that switching depends on interaction state rather than benchmark identity. The controller does not need to know whether a task came from AlphaNLI, Social Chemistry, or closed-ended QA; it reacts to whether the team currently looks prematurely aligned or genuinely resolved.

\subsection{Inclusive Leadership}
\begin{quote}\ttfamily
Input: task $x$, round-$t$ answers $Y_t$ \\
identify current majority and minority positions \\
if a minority objection exists: \\
\hspace*{1em}issue OBJECT to the minority side \\
\hspace*{1em}issue DEFEND to the majority side \\
\hspace*{1em}preserve the objection in the interaction record \\
\hspace*{1em}RECONSIDER before final synthesis \\
else: \\
\hspace*{1em}proceed directly to synthesis
\end{quote}

The defining property is not generic delay, but explicit preservation of minority dissent before convergence.
\end{document}